\documentclass[pdflatex,sn-mathphys-num,iicol]{sn-jnl}

\usepackage{graphicx}%
\usepackage{multirow}%
\usepackage{amsmath,amssymb,amsfonts}%
\usepackage{amsthm}%
\usepackage{mathrsfs}%
\usepackage[title]{appendix}%
\usepackage{xcolor}%
\usepackage{textcomp}%
\usepackage{manyfoot}%
\usepackage{booktabs}%
\usepackage{algorithm}%
\usepackage{algorithmicx}%
\usepackage{algpseudocode}%
\usepackage{listings}%
\usepackage{subcaption}
\usepackage{colortbl}
\usepackage{pifont}
\usepackage{booktabs}
\usepackage[most]{tcolorbox}
\usepackage{dashrule}

\newcommand{\Tref}[1]{Tab.~\ref{#1}}

\newcommand{\Fref}[1]{Fig.~\ref{#1}}
\newcommand{\Sref}[1]{Sec.~\ref{#1}}

\newcommand{\eg}{\textit{e}.\textit{g}.}

\theoremstyle{thmstyleone}%
\theoremstyle{thmstyletwo}%

\theoremstyle{thmstylethree}%

\begin{document}

\title[Article Title]{Towards Reliable Medical LLMs: Benchmarking and Enhancing Confidence Estimation of Large Language Models in Medical Consultation}


\author[1]{\fnm{Zhiyao} \sur{Ren}}\email{zhiyao001@e.ntu.edu.sg}

\author[2]{\fnm{Yibing} \sur{Zhan}}\email{zybjy@mail.ustc.edu.cn}

\author[1]{\fnm{Siyuan} \sur{Liang}}\email{siyuan.liang@ntu.edu.sg}

\author[1]{\fnm{Guozheng} \sur{Ma}}\email{guozheng001@e.ntu.edu.sg}

\author[1]{\fnm{Baosheng} \sur{Yu}}\email{baosheng.yu@ntu.edu.sg}

\author*[1]{\fnm{Dacheng} \sur{Tao}}\email{dacheng.tao@gmail.com}

\affil[1]{\orgname{Nanyang Technological University}, \orgaddress{\city{Singapore}, \country{Singapore}}}

\affil[2]{\orgname{Wuhan University}, \orgaddress{\city{Wuhan}, \country{China}}}

\abstract{
Large-scale language models (LLMs) often offer clinical judgments based on incomplete information, increasing the risk of misdiagnosis.
Existing studies have primarily evaluated confidence in single-turn, static settings, overlooking the coupling between confidence and correctness as clinical evidence accumulates during real consultations, which limits their support for reliable decision-making.
We propose the first benchmark for assessing confidence in multi-turn interaction during realistic medical consultations. 
Our benchmark unifies three types of medical data for open-ended diagnostic generation and introduces an information sufficiency gradient to characterize the confidence–correctness dynamics as evidence increases. 
We implement and compare 27 representative methods on this benchmark; two key insights emerge: (1) medical data amplifies the inherent limitations of token-level and consistency-level confidence methods, and (2) medical reasoning must be evaluated for both diagnostic accuracy and information completeness.
Based on these insights, we present MedConf, an evidence-grounded linguistic self-assessment framework that constructs symptom profiles via retrieval-augmented generation, aligns patient information with supporting, missing, and contradictory relations, and aggregates them into an interpretable confidence estimate through weighted integration.
Across two LLMs and three medical datasets, MedConf consistently outperforms state-of-the-art methods on both AUROC and Pearson correlation coefficient metrics, maintaining stable performance under conditions of information insufficiency and multimorbidity.
These results demonstrate that information adequacy is a key determinant of credible medical confidence modeling, providing a new pathway toward building more reliable and interpretable large medical models.
}




\maketitle

\section{Introduction}

\begin{figure}[!t]
    \centering
    \includegraphics[width=\linewidth]{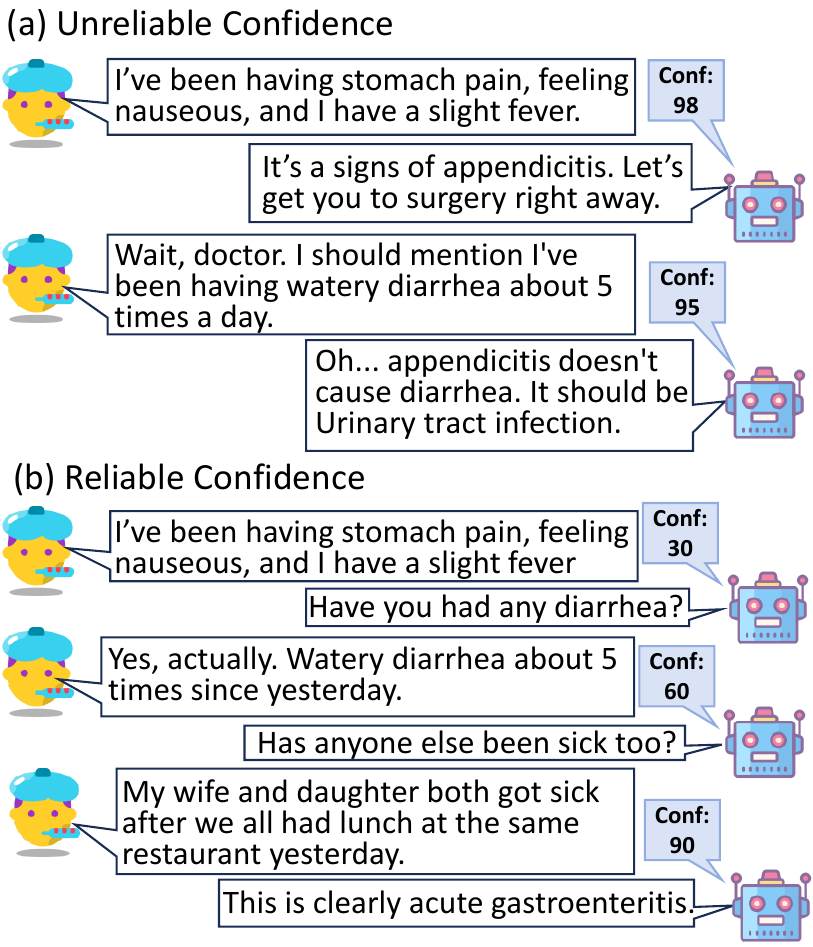}
    \caption{Examples of medical consultation with reliable and unreliable confidence guidance.}
    \label{fig:task}
\end{figure}

LLMs should evaluate their confidence levels in diagnostic conclusions based on the available patient information, similar to how clinicians assess their own confidence during the decision-making process~\cite{meyer2013physicians,ng2007analysis,bhise2018defining}. Such a self-assessment mechanism enables the model to actively acquire more information at low confidence levels and provide diagnostic conclusions at high confidence levels, thereby enhancing the safety of the system in clinical applications.

More broadly, hallucination and unreliable generation remain fundamental obstacles to trustworthy LLM deployment~\cite{ho2024novo,ho2026understanding}. However, the current confidence estimation of LLMs still faces significant challenges: uncalibrated confidence scores may lead to erroneous decisions (\Fref{fig:task}(a)), whereas ideally, confidence should be dynamically adjusted during the interaction process (\Fref{fig:task}(b)).
While studies have begun to investigate the confidence performance of LLMs in the medical domain~\cite{shorinwa2025survey,liu2025uncertainty,savage2025large,atf2025challenge,gu2024probabilistic,omar2024benchmarking,gao2025uncertainty,chen2024uncertainty,ren2026dynamic}, there are still three limitations:, there are still three limitations: (1) Single-round static assessment. Most of the work~\cite{shorinwa2025survey,liu2025uncertainty,savage2025large,atf2025challenge} measures confidence only in a single-round, static scenario, and fails to take into account the dynamic coupling of confidence and diagnostic correctness with the gradual accumulation of clinical evidence.
(2) The task is oversimplified. \citet{gu2024probabilistic} and \citet{omar2024benchmarking} are based on multiple-choice or closed-ended question-and-answer tasks, which are difficult to reflect the context-dependent and complex interactions in real medical consultations;
(3) Limited method coverage. \citet{gao2025uncertainty} and \citet{chen2024uncertainty} have only examined some of the confidence estimation strategies and have not yet systematically compared the applicability of token-level, consistency-level, and self-verbalized methods in medical scenarios.

In this paper, we introduce the first benchmark for assessing the confidence of multi-round interactions in real medical consultations. The benchmark integrates three types of medical data for open-ended diagnostic generation and introduces information sufficiency gradients (1\%, 20\%, 40\%, 60\%, 80\%, 100\%) to characterize the dynamic relationship between confidence and accuracy. We used Pearson and Spearman correlation coefficients~\cite{mukaka2012guide} to assess consistency and measured discriminative power using AUROC~\cite{hanley1982meaning} and AUPRC~\cite{davis2006relationship} on the DDXPlus~\cite{fansi2022ddxplus}, MediTOD~\cite{saley2024meditod}, and MedQA~\cite{jin2021medqa} datasets to compare the performance of the 27 methods.


The evaluation results show that the uncertainty of existing confidence estimation methods on medical data exhibits significant instability and randomness. Some self-verbalized methods achieve better results on DDXPlus and MediTOD but are markedly weaker than token-level and consistency-level methods on MedQA. This inconsistency mainly stems from two reasons. (1) Limitations at the methodological level. The highly specialized terminology and imbalanced label distribution of medical data make token-level methods susceptible to interference, while the latter undermines the stability of consistency-level methods, thereby amplifying their inherent flaws. (2) Domain-specific challenges. Medical diagnosis is inherently uncertain, and the same symptom may correspond to multiple diseases; thus, even if a diagnosis aligns with patient information, it does not necessarily indicate that the information is sufficient or the judgment is reliable. Based on these findings, we propose two takeaways for the design of confidence estimation methods in medical scenarios: (1) confidence estimation methods that rely solely on model output features (e.g., token layer or consistency layer) are highly sensitive to the characteristics of medical data and should be enhanced by incorporating other types of strategies; and (2) confidence assessment in medical tasks should account for both diagnostic accuracy and information completeness rather than only assessing the correctness of answers.

Based on these two insights, we propose MedConf. On one hand, it replaces the reliance on token or consistency signals with self-verbalized evaluation to mitigate its sensitivity to data characteristics at the source. On the other hand, it evaluates information integrity using evidence-based metrics. As shown in \Fref{fig:framework}, the model first generates a preliminary diagnosis based on the available information; then, MedConf retrieves authoritative knowledge related to the diagnosis through retrieval-augmented generation (RAG)~\cite{lewis2020retrieval} and summarizes it to form a symptom spectrum with symptom descriptions and significance; and finally, the LLM identifies the supportive, missing, and contradictory relationships between the information provided by the patient and the symptom spectrum. It weights and aggregates these relationships as interpretable evidence to derive a confidence score for the current diagnosis.

Experimental results demonstrate that MedConf achieves state-of-the-art performance across all datasets and models, exhibiting excellent correlation with accuracy and superior discriminative capability. For example, compared to the 27 baseline methods in our study on DDXPlus with Llama-3.1, MedConf demonstrates average improvements of 0.410 and 0.476 for Pearson and Spearman coefficients, while achieving improvements of 0.129 and 0.124 in AUROC and AUPRC metrics. Furthermore, MedConf demonstrates superior robustness against noisy or irrelevant information and achieves higher accuracy with efficiency when integrated with medical agents. Ablation studies reveal the contribution of different designs in MedConf. Our \textbf{main contributions are}:
\begin{itemize}
    \item We propose the first medical LLMs confidence estimation benchmark that assesses confidence under a multi-turn interaction setting for realistic medical scenarios. Our benchmark utilizes more realistic tasks, comprehensively collects 27 confidence estimation methods, and evaluates their correlation with accuracy and discriminative ability under different levels of patient information. 
    \item Benchmark results reveal that medical data may amplify the limitations of token-level and consistency-level methods that rely on model output features, and medical tasks need to consider uncertainty arising from incomplete input information. To solve this, we propose MedConf, an evidence-based self-verbalized method that considers the supportive, missing, and contradictory relationships between existing information and diagnosis results as evidence to infer a confidence score.
    \item Experimental results show that MedConf achieves state-of-the-art results across all benchmark configurations. Furthermore, MedConf demonstrates robustness to irrelevant information and achieves high diagnostic accuracy and interaction efficiency when integrated into healthcare agents.
\end{itemize}

\section{Related Works}
\subsection{LLMs Confidence Estimation}
\begin{figure}[!t]
    \centering
    \includegraphics[width=\linewidth]{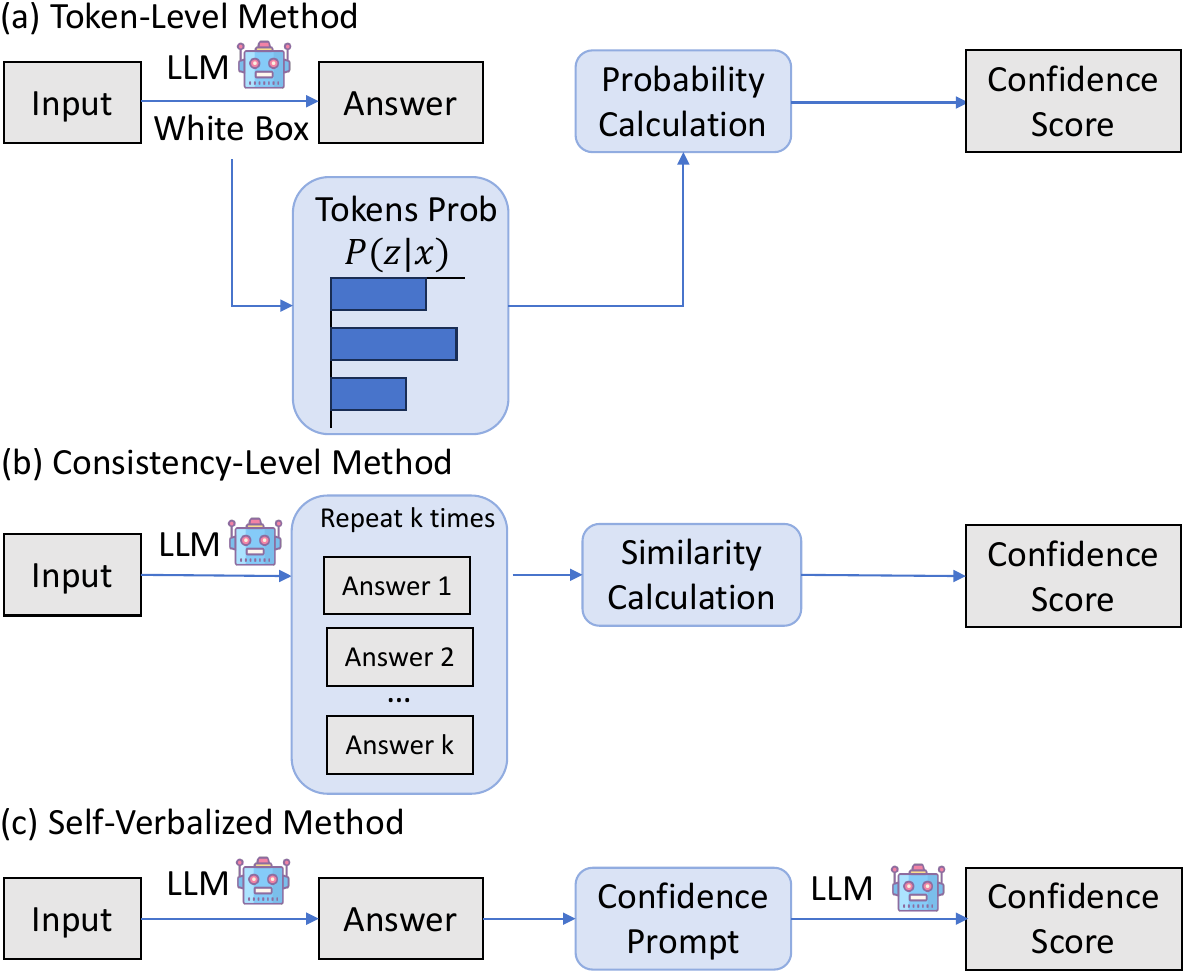}
    \caption{Illustration of the token-level, consistency-level, and self-verbalized method of LLMs confidence estimation.}
    \label{fig:method}
\end{figure}

As illustrated in \Fref{fig:method}, current confidence estimation methods for LLMs can be categorized into three types: token-level methods, consistency-level methods, and self-verbalized methods. In this paper, we collect 27 confidence estimation methods and evaluate them on our proposed medical confidence estimation benchmark.

\textbf{Token-level methods} calculate the confidence of LLMs based on the probability distribution of the generation. These techniques rely on token probability from a single model prediction. \citet{jiang2021can} first propose measuring the confidence score using the predicted probability of the response tokens. \citet{huang2025look} and \citet{manakul2023selfcheckgpt} propose using the negative log-likelihood or entropy of the average or maximum of the response tokens to estimate confidence. Moreover, alternatives such as perplexity~\cite{renout}, mutual information~\cite{takayama2019relevant}, Rényi divergences, and Fisher-Rao distance~\cite{darrin2022rainproof} are widely used to calculate confidence. Using a different approach, Claim Conditional Probability (CCP)~\cite{fadeeva2024fact} decomposes LLM outputs into a set of claims and computes token-level uncertainty from the tokens constituting each claim. Recently, \citet{duan2024shifting} demonstrate that not all tokens contribute equally to the meaning. They propose the tokenSAR method, which re-weights the computation results by evaluating the importance of each token.

\textbf{Consistency-level methods} sample multiple responses to the same query, then utilize the consistency to estimate the confidence score. \citet{manakul2023selfcheckgpt} propose using the percentage of the most consistent samples to evaluate the confidence. \citet{lingenerating} introduce the use of different linear algebra techniques, such as Degree Matrix, Sum of Eigenvalues of the Graph Laplacian, and Eccentricity, to measure confidence. Furthermore, lexical similarity~\cite{manakul2023selfcheckgpt} and semantic similarity~\cite{chen-mueller-2024-quantifying} can also be regarded as essential metrics for consistency assessment. Taking a different approach, some studies integrate token probability into the calculation of consistency measures. Monte Carlo Sequence Entropy~\cite{kuhn2023semantic} generates several sequences via random sampling and computes the resulting entropy. Semantic Entropy~\cite{kuhn2023semantic} clusters the semantically similar generations and then calculates the entropy. Finally, SAR~\cite{duan2024shifting} amplifies the probability of sentences that are more relevant and convincing than others.

\textbf{Self-verbalized methods} query the LLM itself about the confidence of its generation. \citet{lin2022teaching} first demonstrate the LLM's capabilities to provide reasonable confidence. Subsequently, \citet{tian-etal-2023-just} introduce a two-step process, top-k sampling, and chain-of-thought techniques to improve confidence score. P(true)~\cite{kadavath2022language} introduce a different approach by asking the model to validate its answer as true or false and then assigning the confidence score on the probability of true.

\subsection{LLMs Confidence Estimation in Medical Domain}
Confidence estimation effectively improves the reliability and reduces risks of an artificial intelligence system in the medical domain. Existing research has applied confidence estimation to machine learning models~\cite{kompa2021second,kang2021statistical,park2025deep}, vision models~\cite{mehrtash2020confidence,zou2023review,lohr2024towards,abboud2024sparse}, and small language models~\cite{isenegger2019characterizing,peluso2024deep,khandokar2024towards}. However, due to the large-scale parameter characteristics of LLMs, these methods are difficult to transfer to LLM evaluation.
Recent studies have applied existing LLM confidence estimation methods to various medical datasets to investigate their effectiveness on medical data. \citet{savage2025large} and \citet{atf2025challenge} evaluate these methods on Medical QA tasks, while \citet{gu2024probabilistic}, \citet{omar2024benchmarking}, and \citet{savage2025large} focus on medical multiple-choice questions.  \citet{gao2025uncertainty} and \citet{chen2024uncertainty} examine their effectiveness on electronic health records (EHR)-based clinical prediction. However, these benchmarks evaluate confidence only under complete information and overlook how confidence scores evolve as additional relevant information becomes available. Beyond benchmarking, \citet{hu2024uncertainty} introduce confidence scores to enhance inquiry effectiveness, whereas \citet{wu2024uncertainty} and \citet{qin2024enhancing} calibrate medical LLM confidence through Chain-of-Verification and atypical presentations, respectively. In this paper, we propose an evidence-based self-verbalized method that improves confidence estimation for medical datasets.

\begin{table*}[t]
\centering
\caption{Confidence estimation methods implemented in our benchmark.}
\label{tab:method}
\begin{tabular}{lcc}
\toprule
Category & Confidence Estimation Method & Type \\ \midrule
\multirow{10}{*}{Token-level} & Average Sequence Probability (ASP)~\cite{huang2025look} & White-box \\
 & Maximum Sequence Probability (MSP)~\cite{huang2025look} & White-box \\
 & Perplexity~\cite{renout} & White-box \\
 & Entropy~\cite{manakul2023selfcheckgpt} & White-box \\
 & Pointwise Mutual Information (PMI)~\cite{takayama2019relevant} & White-box \\
 & Conditional PMI~\cite{van2022mutual} & White-box \\
 & TokenSAR~\cite{duan2024shifting} & White-box \\
 & Rényi Divergence~\cite{darrin2022rainproof} & White-box \\
 & Fisher-Rao Distance~\cite{darrin2022rainproof} & White-box \\
 & Claim Conditional Probability (CCP)~\cite{fadeeva2024fact} & White-box \\ \midrule
\multirow{13}{*}{Consistency-level} & Percentage of Consistency (PoC)~\cite{manakul2023selfcheckgpt} & Black-box \\
& Lexical Similarity (LexSim)~\cite{manakul2023selfcheckgpt} & Black-box \\
& Semantic Similarity (SemSim) (BERT)~\cite{chen-mueller-2024-quantifying} & Black-box \\
& Semantic Similarity (SemSim) (MedBERT)~\cite{chen-mueller-2024-quantifying} & Black-box \\
& Number of Semantic Sets (NumSet)~\cite{lingenerating} & Black-box \\
& Sum of Eigenvalues of the Graph Laplacian (EigV)~\cite{lingenerating} & Black-box \\
& Degree Matrix (Deg)~\cite{lingenerating} & Black-box \\
& Eccentricity (Ecc)~\cite{lingenerating} & Black-box \\
& Monte Carlo Sequence Entropy (MC-SE)~\cite{kuhn2023semantic} & White-box \\
& Monte Carlo Norm. Seq. Entropy (MC-NSE)~\cite{malininuncertainty} & White-box \\
& Semantic Entropy~\cite{kuhn2023semantic} & White-box \\
& SentenceSAR~\cite{duan2024shifting} & White-box \\
& SAR~\cite{duan2024shifting} & White-box \\ \midrule
\multirow{4}{*}{Self-verbalized} & Confidence Elicitation (CE)~\cite{tian-etal-2023-just} & Black-box \\
& CoT CE~\cite{tian-etal-2023-just} & Black-box \\
& Top-k CE~\cite{tian-etal-2023-just} & Black-box \\
& P(True)~\cite{kadavath2022language} & White-box \\
\bottomrule
\end{tabular}
\end{table*}
\section{Medical Confidence Benchmark}

\subsection{Motivation}
Existing medical LLMs confidence estimation benchmarks~\cite{savage2025large,atf2025challenge,gu2024probabilistic,omar2024benchmarking,gao2025uncertainty,chen2024uncertainty} have three fundamental limitations in their setup. First, current studies only evaluate confidence in a single-round and static scenario, and fail to take into account the dynamic coupling of confidence and diagnostic correctness as the information increases. Second, existing benchmarks only evaluate on simple medical tasks. \citet{savage2025large} and \citet{atf2025challenge} assess confidence performance on simple question-answer task and \citet{gu2024probabilistic} and \citet{omar2024benchmarking} evaluate on multiple-choice questions task. However, open-ended decision-making based on dialogues or reports is more aligned with real-world medical consultations. Finally, existing benchmarks demonstrate insufficient methodological comprehensiveness. \citet{omar2024benchmarking} evaluate only the token probability confidence performance. In contrast, \citet{gu2024probabilistic} and \citet{atf2025challenge} focus on partial token-level and self-verbalized methods. Meanwhile, \citet{savage2025large} includes all three categories of methods but evaluates only 2-3 relatively outdated approaches. Reliable evaluation has likewise become a central concern in broader trustworthy foundation-model research~\cite{zhang2026controllable,ying2026safebench}.

In this section, we introduce a new evaluation benchmark. To conduct testing in scenarios more aligned with real medical practice, we restructure the data format into doctor-patient dialogues or patient reports and transform tasks into open-ended diagnostic generation (In \Sref{sec:data_preparation}). To test the trends of confidence changes as patient information increases, we propose a patient information dividing method and new evaluation metrics (In \Sref{sec:pipeline}). Additionally, we collect and evaluate 27 confidence estimation methods, enhancing the comprehensiveness of benchmark evaluation (In \Sref{sec:methods}).

\subsection{Data Preparation}
\label{sec:data_preparation}
We evaluate the existing confidence estimation methods on three medical datasets: DDXPlus~\cite{fansi2022ddxplus}, MediTOD~\cite{saley2024meditod}, and MedQA~\cite{jin2021medqa}. DDXPlus is a large-scale synthetic dataset containing pathology information, symptoms, and patient antecedents. MediTOD comprises real-world doctor-patient dialogues with annotated complex relationships between dialogue content and corresponding attributes. MedQA is a professional-level multiple-choice question (MCQ) dataset derived from medical licensing examinations (\eg, USMLE) that requires complex medical reasoning and domain knowledge. 

To achieve closer alignment with clinical practice, we preprocess all datasets by transforming them into doctor-patient conversation formats or medical report formats and restructuring them as open-ended diagnosis generation tasks. Specifically, for the DDXPlus dataset, we convert structured data into doctor-patient dialogue format using GPT-4.1~\cite{gpt-4.1}, transforming symptoms into doctor inquiries and converting binary or numerical results into natural language patient responses. For MediTOD, since the data is already in dialogue format, we retain the original structure while filtering to include only effective dialogues based on the purpose and contribution of each conversational turn as labeled in the dataset. For MedQA, we retain its report format and transform the MCQ task into a generation task by removing multiple-choice options and requiring open-ended responses. More details are provided in the supplementary material and Fig. A1.

To ensure adequate patient information for the following confidence estimation, we apply filtering criteria selecting DDXPlus and MediTOD dialogues with more than 10 conversational turns and MedQA cases with more than 10 sentences. The final evaluation dataset comprises 171 DDXPlus instances, 231 MediTOD instances, and 181 MedQA instances.


\subsection{Evaluation Pipeline and Metrics}
\label{sec:pipeline}
In our evaluation, we systematically partition patient information into progressive levels: 1\% (containing only single-turn conversation or one-sentence report), 20\%, 40\%, 60\%, 80\%, and 100\% of the complete case information. We then generate diagnostic predictions using LLMs based on each information level and compute confidence scores for these predictions using different confidence estimation methods. Finally, we assess the accuracy of each generated diagnosis and analyze its relationship with the corresponding confidence scores.

We evaluate confidence estimation performance from two perspectives: 1) Correlation assessment: We employ Pearson and Spearman correlation coefficients to measure the alignment between diagnostic accuracy and confidence scores, determining whether confidence estimates appropriately increase alongside accuracy improvements. 2) Discriminative assessment: We utilize AUROC and AUPRC metrics to evaluate the capability of confidence scores to effectively distinguish between correct and incorrect diagnostic predictions.

\subsection{Evaluated Methods}
\label{sec:methods}
As listed in \Tref{tab:method}, we implement and evaluate 27 widely-used confidence estimation methods, comprising 10 token-level methods, 13 consistency-level methods, and four self-verbalized methods. Accessibility requirements categorize these methods: 11 are black-box methods that rely solely on model generation outputs and apply to both closed-source and open-source models, while 16 are white-box methods that require access to internal model states, such as logits or hidden layer outputs, and are therefore limited to open-source models. For mathematical notations used in this section, please refer to Tab. A1 in the supplementary material.

\noindent\textbf{Token-level methods:} The token-level methods analyze the probability distribution over individual tokens $P(y_l \mid \mathbf{y}_{<l}, \mathbf{x})$, where $\mathbf{x}$ is the input sequence and $\mathbf{y}_{<l}$ is previous generated tokens. A universal formula of token-level methods can be represented as:
\begin{equation}
    C = \varphi\left(\bigoplus_{l=1}^{L} w_l \odot \psi\left(P(y_l \mid \mathbf{y}_{<l}, \mathbf{x})\right)\right)
\end{equation}
where $\psi(\cdot)$ is the token-level transformation function, $\bigoplus_{l=1}^{L}$ is the aggregation operation, $\varphi(\cdot)$ is the final transformation function, $w_l$ is the token weights, and $L$ is the generated sequence length.

For example, Average Sequence Probability (ASP)~\cite{huang2025look} applies identity transformation function ($\psi(P) = P$ and $\varphi(x) = x$) and aggregates tokens probabilities via average ($w_l = \frac{1}{L}$ and $\bigoplus_{l=1}^{L} = \sum_{l=1}^{L}$), while Maximum Sequence Probability (MSP)~\cite{huang2025look} similarly uses identity transformations but aggregates by selecting the maximum token probability ($\bigoplus_{l=1}^{L} = \max_{l=1, ..., L}$). Perplexity~\cite{renout} transforms token probabilities via negative logarithm ($\psi(P) = -\log P$) and applies exponential final transformation ($\varphi(x) = \exp (x)$) with uniform averaging.
Moreover, TokenSAR~\cite{duan2024shifting} proposes that the contributions of different tokens are unequal and utilizes a relevance-based function to change the weight for each token ($w_l = \tilde{R}_T(y_l, \mathbf{y}, \mathbf{x})$). We provide a more detailed discussion on the specifics of these methods and other token-level methods in the supplementary material.

\noindent\textbf{Consistency-level methods:} The consistency-level methods evaluate confidence by assessing the consistency of responses under the same input. This approach involves two steps:

1) Given input $\mathbf{x}$, the LLM generates a response set $\mathcal{Y} = \{\mathbf{y}_1, \mathbf{y}_2, \ldots, \mathbf{y}_K\}$ containing $K$ outputs.

2) Estimate confidence by quantifying the consistency within the set $\mathcal{Y}$.


Consistency-based methods primarily focus on designing different metrics to evaluate consistency. For instance, PoC~\cite{manakul2023selfcheckgpt} calculates the proportion of the most frequent response, while SemSim~\cite{chen-mueller-2024-quantifying} measures confidence by computing the average cosine similarity between the embedding of response pairs. 


On the other hand, several approaches integrate token probability data into consistency measures. Monte Carlo Sequence Entropy (MC-SE)~\cite{kuhn2023semantic} calculates entropy at the sequence level by averaging the negative log-probabilities across multiple generated sequences: 
\begin{equation}
    C_{\text{MC-SE}} = -\frac{1}{K} \sum_{k=1}^{K} \log P(y^{(k)} | \mathbf{x})
\end{equation}
where $P(\mathbf{y}^{(k)} | \mathbf{x}) = \prod_{l=1}^{L_k} P(y_l^{(k)} | \mathbf{y}_{<l}^{(k)}, \mathbf{x})$ represents the probability of the $k$-th generated sequence. Based on this, Semantic Entropy (SE)~\cite{kuhn2023semantic} first clusters the responses into similar groups and then calculates the entropy over these semantic clusters. We provide a more detailed discussion of consistency-based methods in the supplementary material.

\noindent\textbf{Self-verbalized Methods:} The Confidence Elicitation (CE) methods utilize the LLM's reasoning ability to express confidence in natural language~\cite{tian-etal-2023-just}. Given input $\mathbf{x}$ and LLM output $\mathbf{y}$, the model provides a confidence score from 0 to 100 using a specific prompt:
\begin{equation}
    C_{\text{CE}} = f(\text{Prompt}, \mathbf{x}, \mathbf{y})
\end{equation}
where $f(\cdot)$ represents the LLM model. The prompt follows different strategies such as vanilla prompting, Chain-of-Thought reasoning, or Top-k selection, with details provided in the supplementary material and Fig. A2-A4.

Alternatively, the P(True)~\cite{kadavath2022language} method queries the model to validate its answer as true or false. The confidence is computed as the proportion of true responses across $K$ evaluations:
\begin{equation}
    C_{\text{PTrue}} = \frac{1}{K} \sum_{i=1}^{K} \mathbb{I}(y^{(i)} = \text{``True''})
\end{equation}




\begin{figure*}[t]
\centering
\begin{subfigure}{0.45\linewidth}\includegraphics[width=1\linewidth]{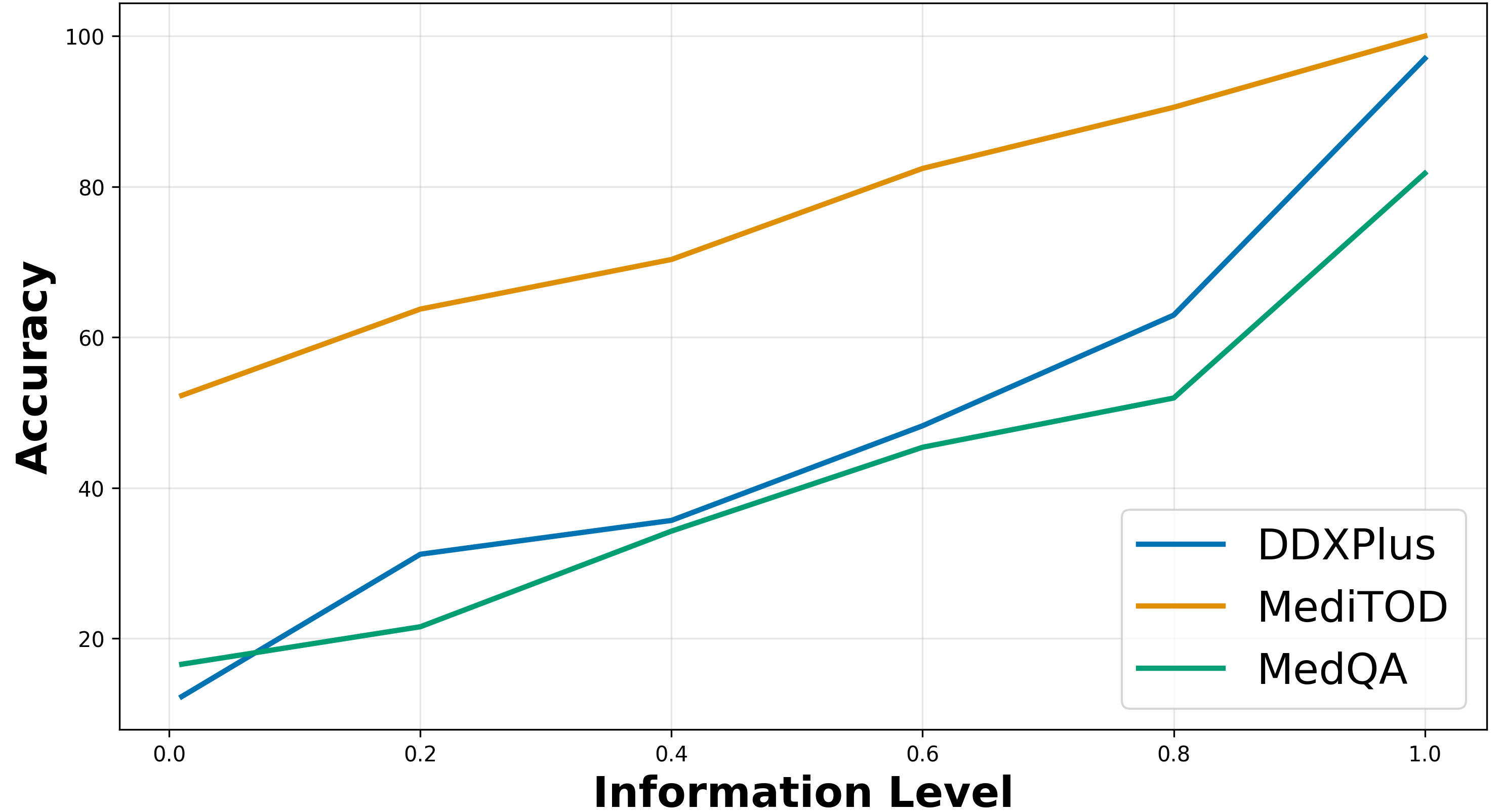}
\caption{Llama-3.1}
\end{subfigure}
\begin{subfigure}{0.45\linewidth}\includegraphics[width=1\linewidth]{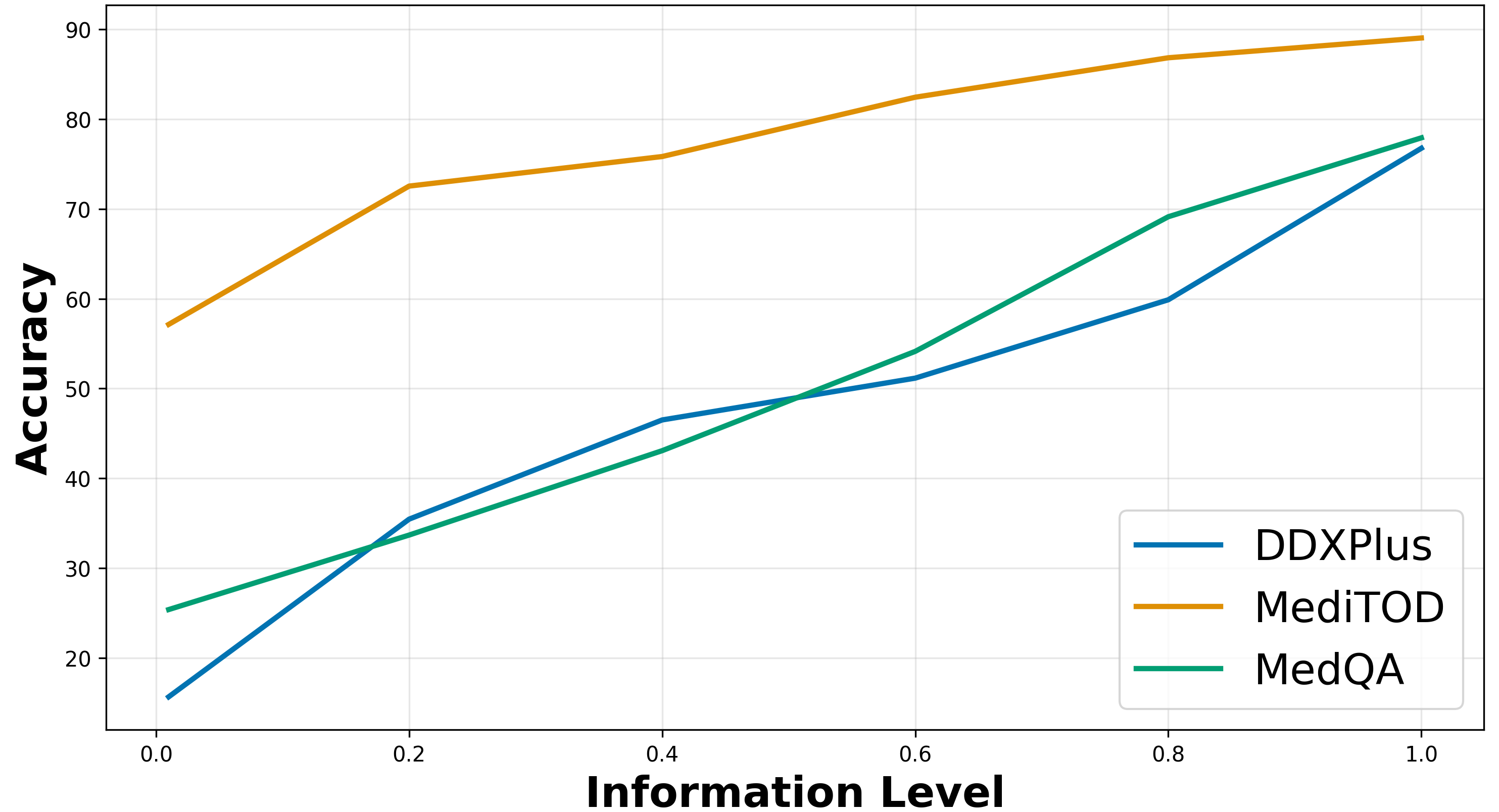}
\caption{GPT-4.1}
\end{subfigure}
\caption{Model performance on medical datasets DDXPlus, MediTOD, and MedQA across information levels for (a) Llama-3.1 and (b) GPT-4.1.}
\label{fig:inform_acc}
\end{figure*}

\definecolor{lightblue}{RGB}{145,173,200}
\definecolor{lightgreen}{RGB}{174,214,207}
\definecolor{lightorange}{RGB}{232,235,196}

\begin{table*}[t]
\centering
\caption{Pearson and Spearman correlation coefficients between accuracy and confidence scores across different confidence estimation methods, models, and datasets. Results show correlation values with p-values in parentheses. Statistically significant results (p$\leq$0.05) are bolded. Token-level methods are colored blue, consistency-level methods are colored green, and self-verbalized methods are colored orange.}
\label{tab:acc_conf}
\begin{subtable}[t]{1\linewidth}
\centering
\caption{Llama-3.1}
\resizebox{1\linewidth}{!}{
\begin{tabular}{lcccccc}
\toprule
\multirow{2}{*}{Method} & \multicolumn{2}{c}{DDXPlus} & \multicolumn{2}{c}{MediTOD} & \multicolumn{2}{c}{MedQA} \\ \cmidrule(lr){2-3} \cmidrule(lr){4-5} \cmidrule(lr){6-7}
& Pearson & Spearman & Pearson & Spearman & Pearson & Spearman \\ \hline
\rowcolor{lightblue}ASP & 0.794 (0.059) & 0.429 (0.397) & 0.306 (0.556) & 0.257 (0.623) & \textbf{0.994 ($\leq$0.05)} & \textbf{0.943 ($\leq$0.05)} \\
\rowcolor{lightblue}MSP & 0.786 (0.064) & \textbf{1.000 ($\leq$0.05)} & 0.807 (0.137) & 0.486 (0.329) & \textbf{0.913 ($\leq$0.05)} & \textbf{0.943 ($\leq$0.05)} \\
\rowcolor{lightblue}Perplexity & 0.508 (0.303) & 0.486 (0.329) & 0.431 (0.596) & 0.086 (0.872) & \textbf{0.978 ($\leq$0.05)} & \textbf{1.000 ($\leq$0.05)} \\
\rowcolor{lightblue}Entropy & 0.14 (0.792) & 0.029 (0.957) & 0.18 (0.733) & 0.143 (0.787) & \textbf{0.905 ($\leq$0.05)} & \textbf{0.829 ($\leq$0.05)} \\
\rowcolor{lightblue}PMI & 0.781 (0.067) & \textbf{0.943 ($\leq$0.05)} & 0.526 (0.283) & 0.714 (0.111) & 0.746 (0.089) & \textbf{0.886 ($\leq$0.05)} \\
\rowcolor{lightblue}Conditional PMI & \textbf{0.829 ($\leq$0.05)} & 0.042 (0.397) & 0.734 (0.097) & \textbf{0.829 ($\leq$0.05)} & \textbf{0.973 ($\leq$0.05)} & \textbf{0.943 ($\leq$0.05)} \\
\rowcolor{lightblue}TokenSAR & 0.515 (0.296) & 0.486 (0.329) & 0.382 (0.226) & 0.257 (0.623) & \textbf{0.981 ($\leq$0.05)} & \textbf{1.000 ($\leq$0.05)} \\
\rowcolor{lightblue}Rényi Divergence & 0.041 (0.939) & 0.029 (0.957) & 0.662 (0.160) & 0.257 (0.623) & \textbf{0.961 ($\leq$0.05)} & 0.771 (0.072) \\
\rowcolor{lightblue}Fisher-Rao Distance & 0.049 (0.927) & 0.086 (0.871) & 0.608 (0.200) & 0.257 (0.623) & \textbf{0.961 ($\leq$0.05)} & \textbf{0.943 ($\leq$0.05)} \\
\rowcolor{lightblue}CCP & \textbf{0.848 ($\leq$0.05)} & 0.943 ($\leq0.05$) & 0.585 (0.223) & 0.771 (0.072) & \textbf{0.958 ($\leq$0.05)} & \textbf{0.943 ($\leq$0.05)} \\ \hline
\rowcolor{lightgreen}PoC & 0.555 (0.253) & 0.2 (0.704) & 0.285 (0.815) & 0.086 (0.872) & \textbf{0.865 ($\leq$0.05)} & \textbf{0.886 ($\leq$0.05)} \\
\rowcolor{lightgreen}Lexical Similarity & 0.320 (0.537) & 0.657 (0.156) & 0.335 (0.516) & 0.657 (0.156) & \textbf{0.917 ($\leq$0.05)} & \textbf{0.829 ($\leq$0.05)} \\
\rowcolor{lightgreen}Semantic Similarity (BERT) & 0.733 (0.098) & 0.429 (0.397) & 0.491 (0.523) & 0.571 (0.156) & \textbf{0.865 ($\leq$0.05)} & \textbf{0.886 ($\leq$0.05)} \\
\rowcolor{lightgreen}Semantic Similarity (MedBERT) & 0.437 (0.386) & 0.2 (0.704) & 0.331 (0.522) & 0.429 (0.397) & 0.265 (0.612) & 0.200 (0.704) \\
\rowcolor{lightgreen}NumSet & 0.443 (0.379) & 0.657 (0.156) & 0.443 (0.379) & 0.657 (0.156) & 0.594 (0.214) & 0.143 (0.787) \\
\rowcolor{lightgreen}EigV & 0.567 (0.24) & 0.029 (0.623) & 0.195 (0.711) & 0.371 (0.468) & \textbf{0.894 ($\leq$0.05)} & 0.600 (0.208) \\
\rowcolor{lightgreen}Deg & 0.655 (0.158) & 0.6 (0.208) & 0.059 (0.911) & 0.600 (0.208) & 0.698 (0.125) & 0.257 (0.623) \\
\rowcolor{lightgreen}Ecc & 0.668 (0.147) & 0.429 (0.397) & 0.668 (0.147) & 0.429 (0.397) & 0.708 (0.116) & 0.429 (0.397) \\
\rowcolor{lightgreen}MC-SE & 0.578 (0.229) & 0.371 (0.468) & 0.579 (0.229) & 0.371 (0.468) & \textbf{0.897 ($\leq$0.05)} & \textbf{0.886 ($\leq$0.05)} \\
\rowcolor{lightgreen}MC-NSE & 0.473 (0.343) & 0.6 (0.208) & 0.473 (0.344) & 0.486 (0.329) & \textbf{0.988 ($\leq$0.05)} & \textbf{1.000 ($\leq$0.05)} \\
\rowcolor{lightgreen}Semantic Entropy & 0.722 (0.105) & \textbf{1.000 ($\leq$0.05)} & 0.510 (0.302) & 0.486 (0.329) & \textbf{0.897 ($\leq$0.05)} & \textbf{1.000 ($\leq$0.05)} \\
\rowcolor{lightgreen}SentenceSAR & 0.326 (0.529) & 0.257 (0.623) & 0.372 (0.467) & 0.257 (0.623) & \textbf{0.981 ($\leq$0.05)} & \textbf{1.000 ($\leq$0.05)} \\
\rowcolor{lightgreen}SAR & 0.347 (0.502) & 0.086 (0.871) & 0.533 (0.280) & 0.486 (0.329) & \textbf{0.897 ($\leq$0.05)} & 0.771 (0.072) \\ \hline
\rowcolor{lightorange}CE & \textbf{0.870 ($\leq$0.05)} & 0.771 (0.072) & \textbf{0.895 ($\leq$0.05)} & \textbf{1.000 ($\leq$0.05)} & \textbf{0.972 ($\leq$0.05)} & \textbf{1.000 ($\leq$0.05)} \\
\rowcolor{lightorange}CoT CE & \textbf{0.907 ($\leq$0.05)} & \textbf{0.943 ($\leq$0.05)} & 0.596 (0.470) & 0.086 (0.623) & \textbf{0.935 ($\leq$0.05)} & \textbf{0.943 ($\leq$0.05)} \\
\rowcolor{lightorange}Top-k CE & \textbf{0.814 ($\leq$0.05)} & \textbf{0.829 ($\leq$0.05)} & 0.690 (0.130) & 0.714 (0.111) & 0.704 (0.119) & 0.771 (0.072) \\
\rowcolor{lightorange}P(True) & 0.364 (0.478) & 0.086 (0.871) & 0.232 (0.558) & 0.143 (0.787) & 0.585 (0.223) & 0.771 (0.072) \\ 
\hline
\end{tabular}
}
\end{subtable}

\vspace{1em}

\begin{subtable}[t]{1\linewidth}
\centering
\caption{GPT-4.1}
\resizebox{1\linewidth}{!}{
\begin{tabular}{lcccccc}
\toprule
\multirow{2}{*}{Method} & \multicolumn{2}{c}{DDXPlus} & \multicolumn{2}{c}{MediTOD} & \multicolumn{2}{c}{MedQA} \\ \cmidrule(lr){2-3} \cmidrule(lr){4-5} \cmidrule(lr){6-7}
& Pearson & Spearman & Pearson & Spearman & Pearson & Spearman \\ \hline
\rowcolor{lightgreen}PoC & 0.105 (0.861) & 0.257 (0.623) & 0.356 (0.489) & 0.600 (0.208) & \textbf{0.834 ($\leq$0.05)} & 0.657 (0.156) \\
\rowcolor{lightgreen}Lexical Similarity & 0.069 (0.897) & 0.371 (0.468) & 0.451 (0.369) & 0.714 (0.111) & \textbf{0.958 ($\leq$0.05)} & \textbf{1.000 ($\leq$0.05)} \\
\rowcolor{lightgreen}Semantic Similarity (BERT) & 0.785 (0.064) & \textbf{0.829 ($\leq$0.05)} & \textbf{0.828 ($\leq$0.05)} & 0.771 (0.072) & \textbf{0.903 ($\leq$0.05)} & \textbf{0.829 ($\leq$0.05)} \\
\rowcolor{lightgreen}Semantic Similarity (MedBERT) & 0.120 (0.821) & 0.257 (0.623) & 0.109 (0.836) & 0.086 (0.872) & \textbf{0.847 ($\leq$0.05)} & 0.771 (0.072) \\
\rowcolor{lightgreen}NumSet & 0.047 (0.930) & 0.086 (0.872) & 0.000 (1.000) & 0.086 (0.872) & 0.804 (0.054) & 0.714 (0.111) \\
\rowcolor{lightgreen}EigV & 0.660 (0.154) & 0.429 (0.397) & \textbf{0.871 ($\leq$0.05)} & \textbf{0.943 ($\leq$0.05)} & \textbf{0.945 ($\leq$0.05)} & \textbf{1.000 ($\leq$0.05)} \\
\rowcolor{lightgreen}Deg & 0.576 (0.231) & 0.657 (0.156) & \textbf{0.873 ($\leq$0.05)} & \textbf{0.943 ($\leq$0.05)} & \textbf{0.900 ($\leq$0.05)} & \textbf{0.829 ($\leq$0.05)} \\
\rowcolor{lightgreen}Ecc & 0.575 (0.233) & 0.600 (0.208) & \textbf{0.873 ($\leq$0.05)} & \textbf{0.943 ($\leq$0.05)} & \textbf{0.906 ($\leq$0.05)} & \textbf{0.886 ($\leq$0.05)} \\ \hline
\rowcolor{lightorange}CE & \textbf{0.989 ($\leq$0.05)} & \textbf{1.000 ($\leq$0.05)} & \textbf{0.981 ($\leq$0.05)} & \textbf{1.000 ($\leq$0.05)} & \textbf{0.989 ($\leq$0.05)} & \textbf{1.000 ($\leq$0.05)} \\
\rowcolor{lightorange}CoT CE & \textbf{0.861 ($\leq$0.05)} & \textbf{0.943 ($\leq$0.05)} & \textbf{0.981 ($\leq$0.05)} & \textbf{1.000 ($\leq$0.05)} & \textbf{0.987 ($\leq$0.05)} & \textbf{1.000 ($\leq$0.05)} \\
\rowcolor{lightorange}Top-k CE & 0.269 (0.591) & 0.429 (0.397) & 0.587 (0.220) & \textbf{0.943 ($\leq$0.05)} & 0.552 (0.256) & 0.486 (0.329) \\
\hline
\end{tabular}
}
\end{subtable}

\end{table*}

\section{Benchmark Results}
In this section, we conduct experiments with two large language models: an open-source model Llama-3.1~\cite{grattafiori2024llama}, and a closed-source model GPT-4.1~\cite{gpt-4.1}. We first report the performance of different methods across various models and dataset combinations, then analyze the shortcomings of existing methods based on experimental results. Finally, we propose insights for designing confidence estimation methods that meet medical requirements.

\subsection{Information-Accuracy Correlation} 
In this part, we examine how diagnostic accuracy varies for Llama-3.1 and GPT-4.1 as patient information is progressively accumulated across different medical datasets. This evaluation establishes the fundamental relationship between information availability and diagnostic performance, providing critical context for understanding the importance of confidence-guided decision making in medical AI systems.

As shown in the results in \Fref{fig:inform_acc}, both models exhibit a consistent and approximately linear increase in diagnostic accuracy with the sequential addition of relevant clinical information. This linear relationship underscores a fundamental principle: making diagnoses based on insufficient patient information inevitably leads to suboptimal outcomes, while comprehensive information gathering significantly enhances diagnostic reliability.

\begin{figure*}[t]
\centering
\begin{subfigure}{1\linewidth}\includegraphics[width=1\linewidth]{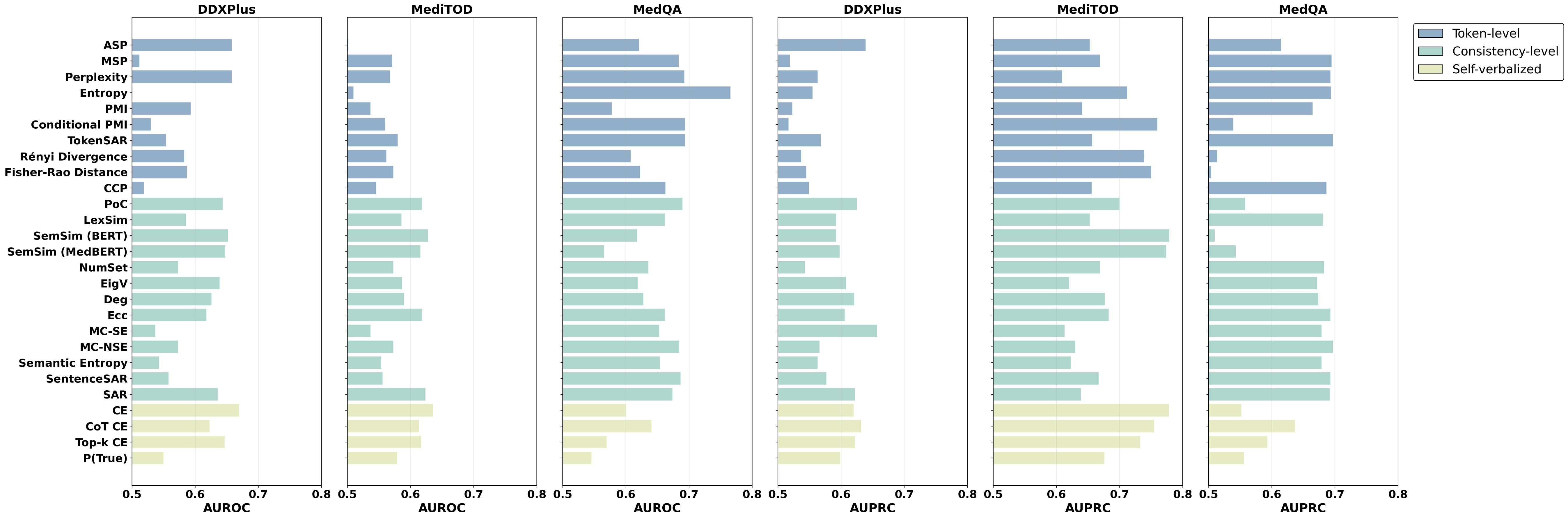}
\caption{Llama-3.1}
\end{subfigure}
\begin{subfigure}{1\linewidth}\includegraphics[width=1\linewidth]{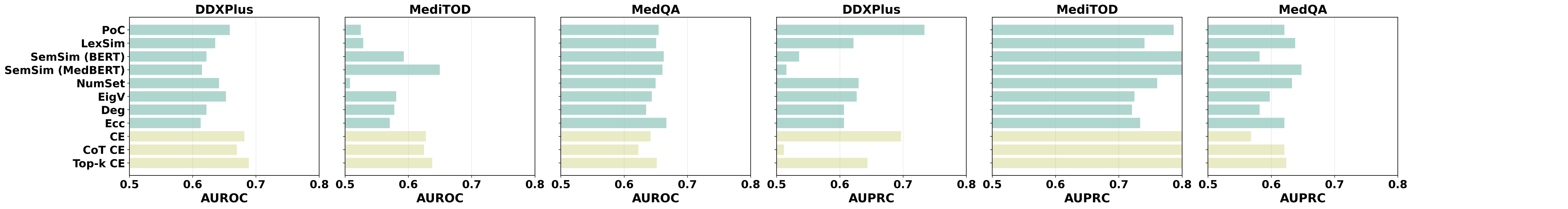}
\caption{GPT-4.1}
\end{subfigure}
\caption{Discriminative performance comparison of confidence estimation methods across models and datasets using AUROC and AUPRC metrics. (a) Results for Llama-3.1 model and (b) Results for GPT-4.1 model. Methods are color-coded by category: token-level methods (blue), consistency-level methods (green), and self-verbalized methods (yellow).}
\label{fig:conf_acc}
\end{figure*}

\subsection{Confidence-Accuracy Correlation}
In this part, we evaluate the correlation between confidence and diagnostic accuracy as patient information increases. The results in \Tref{tab:acc_conf} reveal performance variability across different datasets and models. 
Token-level methods and consistency-level methods exhibit significant sensitivity to datasets and models. For instance, for Llama-3.1, token-level methods Perplexity and TokenSAR demonstrate strong accuracy-confidence alignment on MedQA (Pearson coefficients $>$ 0.961, Spearman coefficients = 1.0), yet show markedly poor consistency on DDXPlus and MediTOD datasets (Pearson coefficients $<$ 0.515, Spearman coefficients $<$ 0.486). Consistency-level methods Eig, Deg, and Ecc methods achieve excellent consistency on MediTOD and MedQA datasets on GPT-4.1, but fail to maintain this performance on DDXPlus with GPT-4.1 or on any datasets with Llama-3.1.

In contrast, self-verbalized methods demonstrate superior robustness across most models and datasets. The CE and CoT CE methods consistently produce confidence scores with statistically significant accuracy correlations across both Llama-3.1 and GPT-4.1 on all datasets. However, exceptions still exist. For example, when applying the CoT CE method to the MediTOD dataset with the Llama-3.1 model, the alignment effect remains limited (Pearson correlation = 0.596, Spearman correlation = 0.086).

\subsection{Discriminative Ability}
In this part, we examine the ability of the confidence score to discriminate between correct and incorrect diagnosis. Based on the results shown in \Fref{fig:conf_acc}, existing confidence estimation methods exhibit two key characteristics in their discriminative ability. First, different methods exhibit varying applicability across datasets. Self-verbalized methods outperform both token-level and consistency-based methods on DDXPlus and MediTOD datasets, while on MedQA, self-verbalized methods demonstrate poor performance and consistency-based methods achieve optimal results. Second,  performance variations exist among methods within the same category. For instance, on DDXPlus using Llama-3.1, the ASP method achieves excellent discriminative ability, whereas the MSP and Entropy methods, despite belonging to the same token-level category, perform substantially worse.

\begin{figure}[t]
\centering
\begin{subfigure}{0.9\linewidth}\includegraphics[width=1\linewidth]{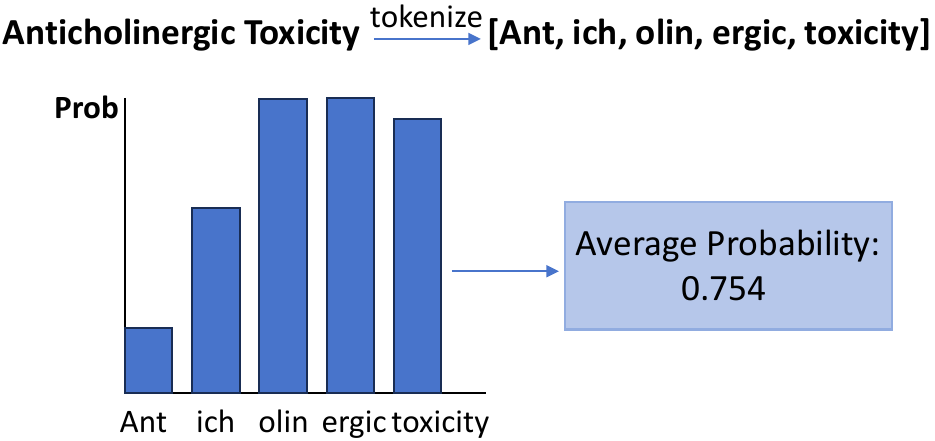}
\caption{Token-level methods limitation}
\end{subfigure}
\begin{subfigure}{0.9\linewidth}\includegraphics[width=1\linewidth]{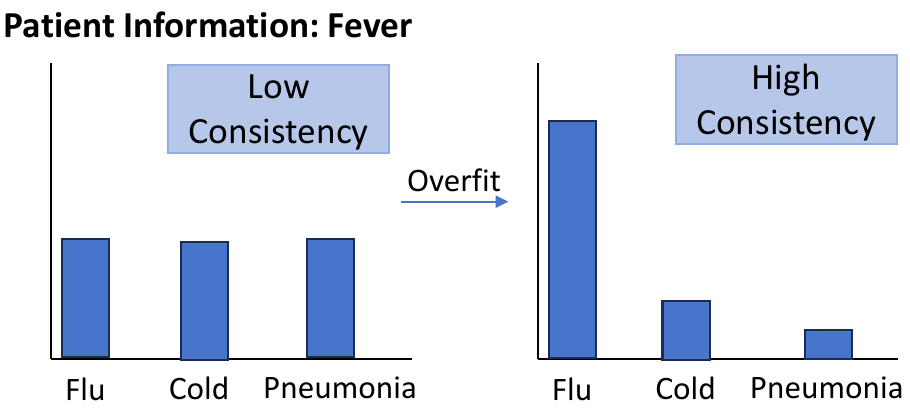}
\caption{Consistency-level methods limitation}
\end{subfigure}
\begin{subfigure}{0.9\linewidth}\includegraphics[width=1\linewidth]{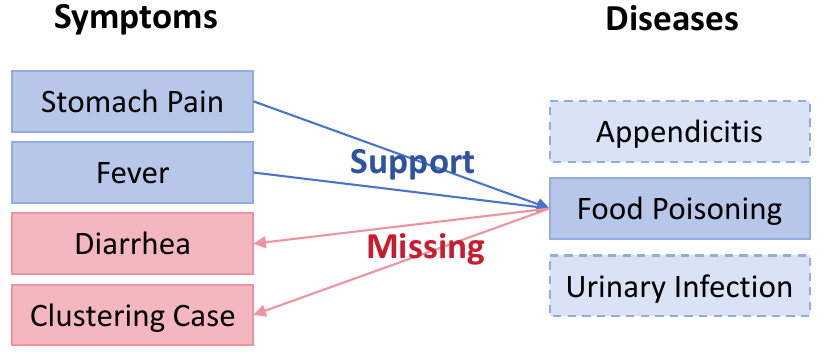}
\caption{Medical domain challenge}
\end{subfigure}
\caption{Example of methodological limitation of (a) token-level methods, (b) consistency-level methods, and (c) medical domain challenge.}
\label{fig:limitation}
\end{figure}

\subsection{Discussion}
Existing methods show certain consistency with accuracy and discriminative ability for correct answers, indicating the potential for applying confidence estimation in the medical domain. However, experimental results reveal that existing methods exhibit sensitivity to both models and datasets. These methods may perform well under specific conditions but substantially underperform when applied to different models or datasets. For example, on Llama-3.1, the Entropy method achieves excellent consistency (Pearson coefficient = 0.905 and Spearman coefficient = 0.829) and discriminative ability (AUROC = 0.766 and AUPRC = 0.694) on the MedQA dataset. In contrast, the same method demonstrates severely limited consistency (Pearson coefficient = 0.14 and Spearman coefficient = 0.029) and discriminative ability (AUROC = 0.501 and AUPRC = 0.555) on the DDXPlus dataset. 

These findings underscore the insufficient reliability of current methods when confronted with diverse medical data, thereby hindering their practical deployment in clinical applications. The challenges of current methods in the medical domain are primarily concentrated in two key areas: 1) methodological limitations, where the unique characteristics of medical data amplify the inherent limitations of existing approaches; 2) domain-specific challenges, where the medical field imposes additional requirements for confidence evaluation.

\subsubsection{Methodological limitations}
Token-level methods exhibit fundamental limitations that they rely heavily on the probability of individual tokens. However, the token logit only captures the model's uncertainty regarding the next token rather than providing an assessment of the reliability of a specific claim~\cite{xiongcan,chen2025mllms,liang2025explaining}. In medical contexts, the complex and lengthy domain-specific terminology magnifies this limitation. As illustrated in \Fref{fig:limitation}(a), which presents an incorrect diagnosis, the \textit{``Anticholinergic Toxicity"} is tokenized into 5 tokens: \textit{``Ant",``ich",``olin",``ergic"}, and \textit{``toxicity"}. Although the model exhibits high uncertainty for the initial token \textit{``Ant"}, once this token is generated, subsequent tokens have very high probabilities. Consequently, when considering the average probability of generating this diagnosis, this erroneous prediction receives a high confidence score of 0.754.

For consistency-level methods, they suffer from a different fundamental issue: consistency reflects only the relative likelihood of an LLM response compared to alternative responses generated by the same model, rather than its likelihood in the real world~\cite{kuhn2023semantic}. In the medical domain, the coexistence of common and rare diseases creates uneven data distributions that amplify the discrepancy between model-generated likelihood and actual clinical likelihood. As the example shown in \Fref{fig:limitation}(b), when patient information contains only non-specific symptoms like fever, conditions such as flu, cold, and pneumonia should theoretically exhibit similar generation probabilities, resulting in appropriately low consistency scores. However, due to the model overfitting toward more frequent conditions like flu, the model generates highly consistent results even for such ambiguous presentations, rendering consistency scores unable to accurately reflect true diagnostic uncertainty.

\subsubsection{Domain-specific Challenge}
Existing confidence estimation methods were originally designed for arithmetic, logic, and symbolic reasoning tasks, which are characterized by comprehensive problem descriptions and unique correct answers~\cite{shorinwa2025survey,liu2025uncertainty}. However, medical diagnostic tasks present fundamentally different challenges: patient-provided information is often incomplete or insufficient, and the relationships between symptoms and diseases are inherently complex and multifaceted. As illustrated in \Fref{fig:limitation}(c), when a patient provides information about stomach pain and fever, the model makes a diagnosis of food poisoning. According to previous task evaluation standards, the current diagnosis contains no errors and should receive high confidence support. However, due to the complex nature of medical conditions, the currently provided information could also correspond to other diseases, such as appendicitis or urinary tract infection, revealing the uncertainty of the current diagnosis. Therefore, in medical contexts, confidence evaluation needs to assess not only whether the conclusion is correct, but also whether the input information is sufficient to support the diagnostic conclusion. However, current methods all lack the ability to evaluate input completeness.

\subsubsection{Medical Confidence Estimation Insights}
Based on existing analysis, we propose two insights for designing medical confidence estimation methods suitable for healthcare applications: \ding{182} Based on our analysis of methodological limitations, methods that rely on model output features (token-level methods and consistency-based methods) are highly sensitive to the characteristics of medical data. To achieve stable performance, alternative strategies that are less affected by medical data should be considered, such as self-verbalized methods based on model reasoning to evaluate confidence scores; \ding{183} Based on our analysis of domain-specific challenges that medical data requires assessing patient information sufficiency, confidence assessment in medical tasks should account for both diagnostic accuracy and information completeness rather than only assessing the correctness of answers.

\begin{figure*}[!t]
    \centering
    \includegraphics[width=\linewidth]{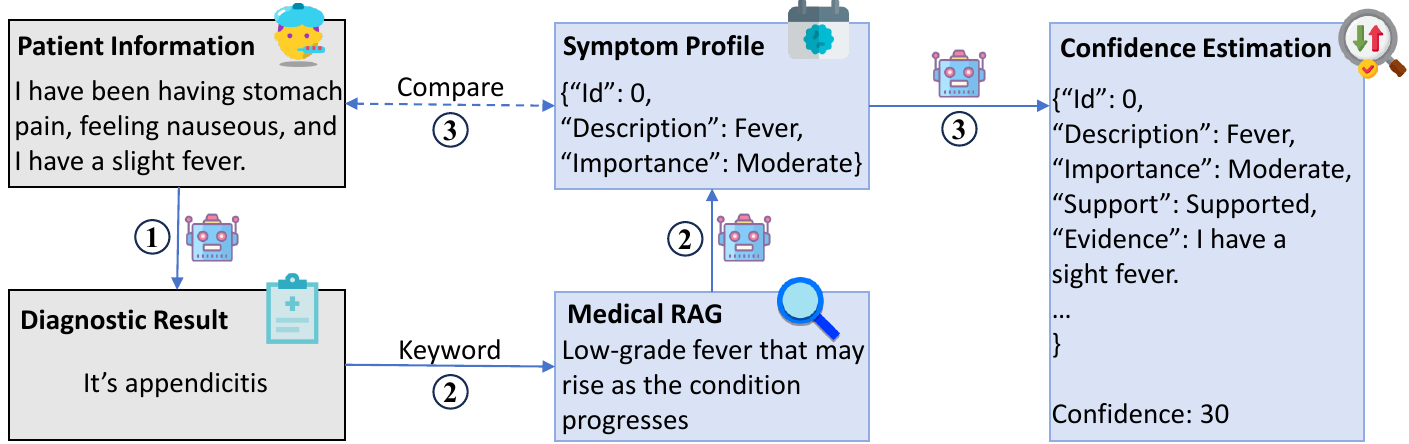}
    \caption{The main proposed MedConf framework. This framework consists three steps: \ding{172} diagnostic generation, \ding{173} symptoms generation, and \ding{174} confidence generation.}
    \label{fig:framework}
\end{figure*}


\section{MedConf}
Following Insight \ding{182}, MedConf does not rely on model output probabilities or distributions, but instead leverages the model's reasoning capabilities through a self-verbalized approach, mitigating its sensitivity to medical data characteristics. According to Insight \ding{183}, MedConf treats the comprehensiveness of patient information as a crucial criterion for confidence assessment. Based on these two insights, we propose MedConf, a diagnostic confidence estimation method based on evidence-driven self-verbalization. The goal of MedConf is to estimate a confidence score $C \in [0,100]$ that reflects the reliability of a diagnosis $d$ generated by an LLM, based on the comprehensiveness and support of available patient evidence. As illustrated in \Fref{fig:framework}, MedConf comprises three sequential steps: diagnostic generation, symptom generation, and confidence generation. We also provide all prompts used in MedConf in the Fig. A5-A7 and a case study for MenConf in the supplementary material and Fig. A9-A13.

\subsection{Diagnostic Generation}
The diagnostic generation step is responsible for producing diagnostic results. Given patient information $I$ and LLM $f$, the model utilizes the diagnosis prompt $\text{Prompt}_d$ to generate diagnostic assessments $d$:
\begin{equation}
    d = f(\text{Prompt}_d, I)
\end{equation}
For instance, the patient information $I$ in \Fref{fig:framework} is ``\textit{I have been having stomach pain, feeling nauseous, and I have a slight fever}'', and the generated diagnosis $d$ is ``\textit{It's appendicitis}''.

\subsection{Symptom Generation}
The symptom generation step operates through a two-phase process to create a comprehensive symptom profile for the predicted diagnosis.

\textbf{Knowledge Retrieval Phase:} The process first utilizes a RAG system to identify and retrieve the relevant medical passages $R$ associated with the generated diagnosis $d$:
\begin{equation}
    R = \text{RAG}(d, \mathcal{D})
\end{equation}
The RAG system enables the LLMs to search relevant literature to integrate domain knowledge in their response. It operates through a three-stage process: First, given the diagnosis $d$, the LLM extracts the primary diagnostic keyword. In the example shown in \Fref{fig:framework}, the keyword is ``\textit{appendicitis}''. Second, the medical corpus $\mathcal{D}$, comprising PubMed, StatPearls, Textbooks, and Wikipedia, is preprocessed by splitting documents into manageable chunks~\cite{xiong2024benchmarking}. Third, we employ BM25~\cite{robertson2009bm25}, a lexical search algorithm, as the retriever to identify the most semantically relevant passages based on keyword matching and term frequency analysis. The number of retrieved passages is determined through empirical validation to balance information comprehensiveness with computational efficiency. In our setting, we choose the 15 most relevant passages as the RAG results. As shown in the example in \Fref{fig:framework}, a passage ``\textit{Low-grade fever that may rise as the condition progresses}'' is retrieved from the corpus.

\textbf{Symptom Profile Generation Phase:} Following the RAG step, the LLM generates structured symptom profiles $S$ to summarize the retrieved medical knowledge:
\begin{equation}
    S = f(\text{Prompt}_s, R, d)
\end{equation}
where $\text{Prompt}_s$ is the symptom profile generation prompt. To ensure the generated content is suitable for subsequent processing, the symptom generation prompt instructs the model to convert retrieved passages into a standardized JSON format with the following schema:
\begin{verbatim}
[
    {
      "id": index number,
      "description": symptom description,
      "importance": Strong|Moderate|Weak
    },
    ...
]
\end{verbatim}
Each symptom entry includes a unique index number, a concise symptom description, and an importance classification that reflects the symptom's diagnostic significance for the given condition. For example, the retrieved passage about fever is converted into ``\textit{\{``id'': 0, ``description'': ``Fever'', ``importance'': ``Moderate''\}}''.

\subsection{Confidence Generation}
The confidence generation process operates through systematic evidence analysis in two distinct phases:

\textbf{Evidence Mapping Phase:} For each symptom $s_i \in S$, the model evaluates its relationship with patient information $I$ using three relationship categories:
\begin{enumerate}
    \item Supported: Patient information explicitly mentions the symptom or closely related clinical manifestations.
    \item Missing: The symptom is absent from patient information despite being expected for the diagnosis, indicating incomplete information.
    \item Contradictory: Patient information explicitly contradicts the expected symptom presentation.
\end{enumerate}
For supported and contradictory cases, relevant statements are extracted from patient information as evidence. This analysis is also generated in a structured format to align with the symptom profiles, ensuring systematic evaluation of each symptom. 

Continuing with our running example from \Fref{fig:framework}, the symptom ``\textit{fever}'' is compared against the patient statement ``\textit{I have a slight fever''}. Since this represents explicit symptom mention, it is categorized as \textit{Supported} with the evidence text.

\textbf{Score Aggregation Phase:} The model summarizes evidence from all symptoms to produce the final confidence score. The model is guided to consider both the proportion of supported, missing, and contradictory symptoms and their relative importance weights, with higher confidence assigned when critical symptoms are well-supported and fewer missing or contradictions are present. The overall confidence generation process is formulated as:
\begin{equation}
    C_{\text{MedConf}} = f(\text{Prompt}_c, S, I, d)
\end{equation}
where $\text{Prompt}_c$ is the prompt for confidence score generation. For our example in \Fref{fig:framework}, although the symptom fever is supported by evidence, there are still other decisive symptoms that are not supported and are missing. Therefore, the model inference is that the final confidence score is 30.


\section{Experimental Results}
In this section, we conduct comprehensive evaluations of MedConf. First, we compare MedConf against existing confidence estimation methods to assess its performance across various model-dataset combinations. Second, we validate MedConf's robustness against irrelevant information interference, which is a critical requirement for reliable deployment in clinical settings where extraneous information may be present. Third, we evaluate the effectiveness of integrating MedConf with healthcare agents to assess its practical utility in interactive diagnostic scenarios. Finally, we conduct ablation studies to quantify the contribution of each key design component within MedConf.

\subsection{Comparison with Existing Methods}
We systematically evaluate MedConf against the best-performing baseline methods of token-level, consistency-level, and self-verbalized categories across all model-dataset combinations. As demonstrated in \Tref{tab:all_results}, MedConf exhibits consistent and substantial superiority across every evaluated configuration, establishing its robustness across different model architectures and medical datasets.

\textbf{Confidence-Accuracy Correlation:} For confidence-accuracy correlation, MedConf demonstrates exceptional performance with Pearson correlation coefficients ranging from 0.936 to 0.989. It achieves either the highest correlation or performs comparably to the best baseline across all conditions. This superior alignment indicates that MedConf's confidence scores are more calibrated and trustworthy indicators of prediction accuracy.

\textbf{Discriminative Ability:} MedConf achieves notable improvements in discriminative capability, with an average AUROC enhancement of 0.056 compared to the best-performing baseline methods across all configurations. Specifically, MedConf attains AUROC scores ranging from 0.672 to 0.796, consistently outperforming competing approaches by margins of 0.030 to 0.126 depending on the model-dataset combination. These improvements demonstrate MedConf's superior ability to distinguish between correct and incorrect predictions, establishing it as a more reliable confidence estimation method for medical AI applications.

\textbf{Generalizability:} The consistent performance improvements across three distinct medical datasets (DDXPlus , MediTOD, and MedQA) and two different model architectures (Llama-3.1 and GPT-4.1) demonstrate MedConf's broad applicability and reliability within the medical domain. These results collectively establish that MedConf generates more reliable and well-calibrated confidence scores, providing enhanced stability and trustworthiness compared to existing confidence estimation methods in medical LLMs applications.

\begin{table*}[t]
\centering
\caption{Performance comparison of MedConf against the best-performing token-level, consistency-level, and self-verbalized methods across Llama-3.1 and GPT-4.1 on DDXPlus, MediTOD, and MedQA.}
\label{tab:all_results}
\begin{subtable}{0.32\textwidth}
\centering
\caption{Llama-3.1 and DDXPlus}
\resizebox{!}{0.7cm}{
\begin{tabular}{lccccc}
\toprule
Method & Type & AUROC & Pearson \\
\midrule
MedConf & Self-Verbalized & \textbf{0.722} & \textbf{0.968} \\
PMI & Token-Level & 0.593 & 0.829 \\
SemSim (BERT) & Consistency-Level & 0.652 & 0.733 \\
CE & Self-Verbalized & 0.670 & 0.870 \\
\bottomrule
\end{tabular}
}
\end{subtable}
\hfill
\begin{subtable}{0.32\textwidth}
\centering
\caption{Llama-3.1 and MediTOD}
\resizebox{!}{0.7cm}{
\begin{tabular}{lccccc}
\toprule
Method & Type & AUROC & Pearson \\
\midrule
MedConf & Self-Verbalized & \textbf{0.687} & \textbf{0.943} \\
Conditional PMI & Token-Level & 0.560 & 0.734 \\
SemSim (BERT) & Consistency-Level & 0.628 & 0.491 \\
CE & Self-Verbalized & 0.636 & 0.895 \\
\bottomrule
\end{tabular}
}
\end{subtable}
\hfill
\begin{subtable}{0.32\textwidth}
\centering
\caption{Llama-3.1 and MedQA}
\resizebox{!}{0.7cm}{
\begin{tabular}{lccccc}
\toprule
Method & Type & AUROC & Pearson \\
\midrule
MedConf & Self-Verbalized & \textbf{0.796} & 0.979 \\
Entropy & Token-Level & 0.766 & 0.905 \\
MC-NSE & Consistency-Level & 0.685 & \textbf{0.988} \\
CoT CE & Self-verbalized & 0.641 & 0.935 \\
\bottomrule
\end{tabular}
}
\end{subtable}
\hfill
\begin{subtable}{0.32\textwidth}
\centering
\caption{GPT-4.1 and DDXPlus}
\resizebox{!}{0.6cm}{
\begin{tabular}{lccccc}
\toprule
Method & Type & AUROC & Pearson \\
\midrule
MedConf & Self-Verbalized & \textbf{0.795} & 0.936 \\
SemSim (BERT) & Consistency-Level & 0.622 & 0.785 \\
CE & Self-Verbalized & 0.682 & \textbf{0.989} \\
\bottomrule
\end{tabular}
}
\end{subtable}
\hfill
\begin{subtable}{0.32\textwidth}
\centering
\caption{GPT-4.1 and MediTOD}
\resizebox{!}{0.6cm}{
\begin{tabular}{lccccc}
\toprule
Method & Type & AUROC & Pearson \\
\midrule
MedConf & Self-Verbalized & \textbf{0.672} & \textbf{0.981} \\
SemSim (BERT) & Consistency-Level & 0.593 & 0.829 \\ 
CE & Self-Verbalized & 0.628 & \textbf{0.981} \\
\bottomrule
\end{tabular}
}
\end{subtable}
\hfill
\begin{subtable}{0.32\textwidth}
\centering
\caption{GPT-4.1 and MedQA}
\resizebox{!}{0.6cm}{
\begin{tabular}{lccccc}
\toprule
Method & Type & AUROC & Pearson \\
\midrule
MedConf & Self-Verbalized & \textbf{0.690} & \textbf{0.989}  \\
Ecc & Consistency-Level & 0.667 & 0.906 \\
CoT CE & Self-Verbalized & 0.642 & 0.987 \\
\bottomrule
\end{tabular}
}
\end{subtable}
\end{table*}

\begin{table}[t]
\centering
\caption{Robustness evaluation of confidence estimation methods to irrelevant information. Coefficient of variation (CV) measures performance stability of mean and sample level.}
\label{tab:useless_information}
\setlength{\tabcolsep}{3pt}
\begin{tabular}{lcccc}
\toprule
Method & Type & CV group & CV sample \\ \midrule
MedConf & Self-Verbalized & 2.343 & 12.401 \\
ASP & Token-Level & 2.580 & 15.258 \\
PoC & Consistency-Level & 4.858 & 24.325 \\
CE & Self-Verbalized & 11.655 & 47.575 \\ 
\bottomrule
\end{tabular}
\end{table}

\subsection{Robustness to Irrelevant Information}
We conduct a robustness analysis comparing MedConf against representative baseline methods from different confidence estimation categories: ASP (token-level), PoC (consistency-level), and CE (self-verbalized). This evaluation is performed using Llama-3.1 on the DDXPlus dataset to assess each method's resistance to irrelevant information interference, a critical factor for reliable clinical deployment.

\textbf{Experimental Setup:} To simulate realistic clinical scenarios where additional but non-diagnostic information may be present, we randomly select 1-2 conversational turns from doctor-patient dialogues and employ GPT-4.1 to generate semantically equivalent paraphrased content with varied linguistic expressions. This paraphrased content is appended to the model input, creating controlled conditions where the added information provides no additional diagnostic value while potentially introducing linguistic variations that could affect confidence estimation stability.

We evaluate confidence scores across three distinct irrelevant information conditions using identical random seeds to ensure fair comparison. Robustness is quantified using the Coefficient of Variation (CV), where CV group measures variability among mean confidence scores across the three experimental groups, and CV sample calculates the average CV across individual sample instances. Lower CV values indicate greater robustness and stability.

\textbf{Results and Analysis:} As demonstrated in \Tref{tab:useless_information}, baseline methods exhibit substantial vulnerability to irrelevant information interference. The CE method shows particularly poor stability with CV values of 11.655 (group-level) and 47.575 (sample-level), likely attributable to the inherent sensitivity of self-verbalized approaches to prompt variations and linguistic nuances. In contrast, MedConf demonstrates exceptional robustness with significantly lower variability scores (CV group: 2.343, CV sample: 12.401). This represents substantial improvements of 0.237 and 2.857 over ASP, 2.515 and 11.924 over PoC, and remarkable improvements of 9.312 and 35.174 over the CE method. 

MedConf's superior robustness stems from its fundamental design principle: MedConf is based on structured patient information analysis rather than surface-level linguistic features. Since semantically equivalent but linguistically varied content does not alter the underlying patient information, MedConf maintains stable confidence scores regardless of presentation variations. These results establish MedConf's superior predictability and robustness across varying input scenarios—essential characteristics for reliable deployment in medical AI applications where consistency and stability are important for supporting critical clinical decision-making processes.

\begin{table}[t]
\centering
\caption{Performance of confidence estimation methods integrated into healthcare agents. $\#$ Utterances represents the average utterances in a consultation, and accuracy indicates diagnostic accuracy.}
\label{tab:agent}
\setlength{\tabcolsep}{3pt}
\begin{tabular}{lcccc}
\toprule
Method & Type & $\#$ Utterances & Accuracy (\%) \\ \midrule
MedConf & Self-Verbalized & 17.84 & 36 \\
ASP & Token-Level & 5.20 & 24 \\
PoC & Consistency-Level & 18.00 & 30 \\
CE & Self-Verbalized & 25.04 & 36 \\
\bottomrule
\end{tabular}
\end{table}

\begin{table*}[t]
\centering
\caption{The results of ablation studies.}
\label{tab:ablation}
\begin{tabular}{lcccc}
\toprule
Method & AUROC & AUPRC & Pearson & Spearman \\ \midrule
MedConf & 0.772 & 0.708 & 0.968 & 0.943 \\ \midrule
\quad w/o Symptom Profile & 0.657 & 0.645 & 0.786 & 0.657 \\
\quad w/o RAG & 0.670 & 0.588 & 0.939 & 0.771 \\
\quad w/o Structure Format & 0.688 & 0.692 & 0.907 & 0.943 \\
\quad w/o Importance Level & 0.639 & 0.601 & 0.959 & 0.943 \\
\bottomrule
\end{tabular}
\end{table*}

\subsection{Integration with Healthcare Agent}
Recent studies have also highlighted the importance of reliability and safety evaluation at the agent and interaction levels~\cite{liu2025agentsafe,liang2025safemobile,liu2025roboview}. We integrate MedConf and three representative baseline methods (ASP, PoC, and CE) into healthcare agents to evaluate their effectiveness in improving diagnostic performance within interactive clinical scenarios. This evaluation assesses how different confidence estimation methods impact both diagnostic accuracy and interaction efficiency in practical healthcare applications.

\textbf{Experimental Setup:} Following the setup of \citet{ren2024healthcare}, we implement healthcare agents that employ confidence score-based decision making rather than model planning approaches. The agent determines the next action as either patient inquiry or diagnosis based on confidence score thresholds: the agent continues patient inquiry when confidence scores fall below a predetermined threshold and proceeds to diagnosis when scores exceed this threshold. We deploy these agents in simulated clinical consultations with virtual patients, systematically evaluating each confidence estimation method's impact on diagnostic performance. The effectiveness is measured through two key metrics: diagnostic accuracy (percentage of correct diagnoses) and average dialogue length (measured in utterances per consultation), which together capture the critical trade-off between diagnostic quality and resource efficiency in clinical settings.

\textbf{Results and Analysis:} As shown in \Tref{tab:agent}, the confidence estimation methods exhibit distinct behavioral patterns that directly impact both diagnostic accuracy and interaction efficiency. The ASP method demonstrates a tendency to generate inappropriately high confidence scores during early interaction stages, resulting in premature diagnostic decisions with only 5.20 average utterances per consultation. This aggressive early termination leads to suboptimal diagnostic accuracy of 24\%, indicating insufficient information gathering for reliable diagnosis. Conversely, the CE method produces high confidence scores only when substantial patient information becomes available, leading to extended dialogues averaging 25.04 utterances per consultation. While this conservative approach enables CE to achieve higher diagnostic accuracy (36\%) compared to ASP, it requires significantly more computational resources and consultation time due to prolonged interactions, potentially limiting practical applicability in resource-constrained clinical settings. The PoC method demonstrates intermediate behavior with 18.00 average utterances and 30\% accuracy, showing moderate performance across both efficiency and accuracy metrics but failing to optimize either dimension effectively.

MedConf achieves superior performance by striking an optimal balance between diagnostic accuracy and interaction efficiency.   With 17.84 average utterances per consultation, MedConf maintains computational efficiency comparable to PoC while achieving the highest diagnostic accuracy of 36\%—matching CE's performance with significantly reduced resource requirements.   These results demonstrate that MedConf's well-calibrated confidence scores enable more informed decision-making in healthcare agents, leading to appropriately timed diagnostic decisions that maximize accuracy while minimizing unnecessary prolonged interactions.

\subsection{Ablation Studies}
In this section, we conduct systematic ablation experiments to investigate the contribution of key design components within MedConf. These experiments isolate individual components to quantify their specific impact on confidence estimation performance. The experimental results are presented in \Tref{tab:ablation}, evaluated using Llama-3.1 on the DDXPlus dataset across multiple performance metrics, including AUROC, AUPRC, Pearson correlation, and Spearman correlation.

\textbf{Two-Step Reasoning Process:} We first evaluate the impact of MedConf's symptom-confidence two-step reasoning framework by removing the intermediate symptom profile generation step. In this ablation variant (w/o Symptom Profile), we directly generate confidence scores using RAG-retrieved results, bypassing the symptom profile generation phase. The results demonstrate substantial performance degradation across all metrics, with AUROC declining from 0.772 to 0.657, AUPRC dropping from 0.708 to 0.645, and correlation coefficients decreasing significantly (Pearson: 0.968 to 0.786, Spearman: 0.943 to 0.657). These findings reveal that models struggle to effectively utilize raw RAG retrieval results as the direct foundation for confidence reasoning. The intermediate symptom profile generation step addresses this limitation by transforming fragmented retrieval information into coherent, structured representations that are more amenable to model interpretation, thereby substantially enhancing the reasoning capability for accurate confidence score generation.

\textbf{RAG Integration:} Secondly, we evaluate MedConf's performance without external knowledge retrieval (w/o RAG) to assess the necessity of incorporating medical knowledge bases. Removing RAG causes substantial performance degradation, with AUROC declining from 0.772 to 0.670, AUPRC decreasing from 0.708 to 0.588, Pearson correlation dropping from 0.968 to 0.939, and Spearman correlation decreasing from 0.943 to 0.771. These findings indicate that external knowledge retrieval is crucial for achieving optimal discriminative capability in medical confidence estimation, as it provides essential domain-specific context that enhances the model's ability to assess prediction reliability accurately.

\textbf{Structured Format and Importance-Level Annotations:} Finally, we examine the contribution of structured formatting and importance-level annotations during symptom profile generation through two separate ablations. Removing structured format (w/o Structure Format) results in moderate performance decline (AUROC from 0.772 to 0.688, AUPRC from 0.708 to 0.692), while eliminating importance level annotations (w/o Importance Level) causes more substantial degradation in discriminative performance (AUROC from 0.772 to 0.639, AUPRC from 0.708 to 0.601). The experimental results indicate that the structured format improves the interpretability and recognition capability of symptom profiles, preventing information loss during subsequent reasoning processes.  More critically, importance level annotations provide essential weighting information that enables the model to appropriately consider each symptom's differential impact on confidence estimation based on clinical significance. 

\section{Conclusion and Future Work}
Reliable confidence estimation is essential for medical LLMs to prevent misdiagnosis caused by premature conclusions. In this work, we introduce the first comprehensive benchmark that evaluates 27 confidence estimation methods across three medical datasets under varying levels of information, assessing both their correlation with accuracy and their discriminative capability. Our results show that existing methods suffer from instability in medical tasks, driven by methodological limitations and the unique requirement to assess not only diagnostic accuracy but also the completeness of patient information. To address these gaps, we propose MedConf, a diagnostic evidence–based self-verbalized method that leverages retrieval-augmented generation to analyze the relationships between patient information and symptom profiles. MedConf achieves state-of-the-art performance, demonstrating superior accuracy correlation, discriminative power, and robustness, while preserving accuracy and efficiency when deployed within healthcare agents.

\textbf{Limitation and Future Work:} While the benchmark and method presented in this paper are promising, several limitations remain to be addressed in future work. First, our evaluation focuses solely on diagnostic tasks. Confidence estimation, however, is also important for other medical applications such as medical report generation and clinical note summarization, which introduce different challenges. These tasks are beyond the current scope, but we plan to extend the benchmark to cover a broader range of medical applications.
Second, although our MedConf method achieves state-of-the-art performance, its multi-step reasoning process introduces additional computational overhead and latency compared to simpler token-level approaches. This may limit real-time deployment in some resource-constrained clinical settings. To address this, we plan to investigate efficiency optimization techniques, including caching frequently accessed medical knowledge and developing lightweight symptom profile generation models, to reduce inference time while maintaining performance. 
Finally, all experiments in this paper are conducted in simulated environments using benchmark datasets rather than in real clinical settings. While these datasets reflect real-world data and provide a practical alternative to clinical testing at this stage—helping to reduce the burden on medical professionals and patients as well as the time required—they cannot fully capture the complexities of real-world clinical practice. As future work, we aim to conduct prospective clinical trials where healthcare professionals interact directly with confidence-aware AI systems to assess their practical utility and the interpretability of confidence scores.

\section*{Author contribution}
Z.R., Y.Z., B.Y., and G.M. conceived and designed this study. Z.R. developed code and conducted experiments. Z.R., Y.Z., B.Y., S.L., and D.T. contributed to write the manuscript. Y.Z., B.Y, S.L., and D.T. supervised this paper. All authors have read and approved the manuscript.

\section*{Acknowledgements}
This study received no funding.

\section*{Funding}
This study received no funding.

\section*{Competing interests}
The authors declare no competing interests.

\section*{Data availability}
The DDXPlus, MediTOD, and MedQA datasets are publicly available.

\section*{Code availability}
The underlying code for this study is not publicly available but may be made available to qualified researchers on reasonable request from the corresponding author.

\bibliography{sn-bibliography}

@article{meyer2013physicians,
  title={Physicians’ diagnostic accuracy, confidence, and resource requests: a vignette study},
  author={Meyer, Ashley ND and Payne, Velma L and Meeks, Derek W and Rao, Radha and Singh, Hardeep},
  journal={JAMA internal medicine},
  volume={173},
  number={21},
  pages={1952--1958},
  year={2013},
  publisher={American Medical Association}
}

@article{ng2007analysis,
  title={Analysis of diagnostic confidence and diagnostic accuracy: a unified framework},
  author={Ng, CS and Palmer, CR},
  journal={The British journal of radiology},
  volume={80},
  number={951},
  pages={152--160},
  year={2007},
  publisher={British Institute of Radiology}
}

@article{bhise2018defining,
  title={Defining and measuring diagnostic uncertainty in medicine: a systematic review},
  author={Bhise, Viraj and Rajan, Suja S and Sittig, Dean F and Morgan, Robert O and Chaudhary, Pooja and Singh, Hardeep},
  journal={Journal of general internal medicine},
  volume={33},
  number={1},
  pages={103--115},
  year={2018},
  publisher={Springer}
}

@article{shorinwa2025survey,
  title={A survey on uncertainty quantification of large language models: Taxonomy, open research challenges, and future directions},
  author={Shorinwa, Ola and Mei, Zhiting and Lidard, Justin and Ren, Allen Z and Majumdar, Anirudha},
  journal={ACM Computing Surveys},
  year={2025},
  publisher={ACM New York, NY}
}

@inproceedings{liu2025uncertainty,
  title={Uncertainty quantification and confidence calibration in large language models: A survey},
  author={Liu, Xiaoou and Chen, Tiejin and Da, Longchao and Chen, Chacha and Lin, Zhen and Wei, Hua},
  booktitle={Proceedings of the 31st ACM SIGKDD Conference on Knowledge Discovery and Data Mining V. 2},
  pages={6107--6117},
  year={2025}
}

@article{jiang2021can,
  title={How can we know when language models know? on the calibration of language models for question answering},
  author={Jiang, Zhengbao and Araki, Jun and Ding, Haibo and Neubig, Graham},
  journal={Transactions of the Association for Computational Linguistics},
  volume={9},
  pages={962--977},
  year={2021},
  publisher={MIT Press One Rogers Street, Cambridge, MA 02142-1209, USA journals-info~…}
}

@article{huang2025look,
  title={Look before you leap: An exploratory study of uncertainty analysis for large language models},
  author={Huang, Yuheng and Song, Jiayang and Wang, Zhijie and Zhao, Shengming and Chen, Huaming and Juefei-Xu, Felix and Ma, Lei},
  journal={IEEE Transactions on Software Engineering},
  year={2025},
  publisher={IEEE}
}

@inproceedings{manakul2023selfcheckgpt,
  title={SelfCheckGPT: Zero-Resource Black-Box Hallucination Detection for Generative Large Language Models},
  author={Manakul, Potsawee and Liusie, Adian and Gales, Mark},
  booktitle={Proceedings of the 2023 Conference on Empirical Methods in Natural Language Processing},
  pages={9004--9017},
  year={2023}
}

@article{kuhn2023semantic,
  title={Semantic uncertainty: Linguistic invariances for uncertainty estimation in natural language generation},
  author={Kuhn, Lorenz and Gal, Yarin and Farquhar, Sebastian},
  journal={arXiv preprint arXiv:2302.09664},
  year={2023}
}

@inproceedings{duan2024shifting,
  title={Shifting Attention to Relevance: Towards the Predictive Uncertainty Quantification of Free-Form Large Language Models},
  author={Duan, Jinhao and Cheng, Hao and Wang, Shiqi and Zavalny, Alex and Wang, Chenan and Xu, Renjing and Kailkhura, Bhavya and Xu, Kaidi},
  booktitle={Proceedings of the 62nd Annual Meeting of the Association for Computational Linguistics (Volume 1: Long Papers)},
  pages={5050--5063},
  year={2024}
}

@article{kadavath2022language,
  title={Language models (mostly) know what they know},
  author={Kadavath, Saurav and Conerly, Tom and Askell, Amanda and Henighan, Tom and Drain, Dawn and Perez, Ethan and Schiefer, Nicholas and Hatfield-Dodds, Zac and DasSarma, Nova and Tran-Johnson, Eli and others},
  journal={arXiv preprint arXiv:2207.05221},
  year={2022}
}

@inproceedings{tian-etal-2023-just,
    title = "Just Ask for Calibration: Strategies for Eliciting Calibrated Confidence Scores from Language Models Fine-Tuned with Human Feedback",
    author = "Tian, Katherine  and
      Mitchell, Eric  and
      Zhou, Allan  and
      Sharma, Archit  and
      Rafailov, Rafael  and
      Yao, Huaxiu  and
      Finn, Chelsea  and
      Manning, Christopher",
    editor = "Bouamor, Houda  and
      Pino, Juan  and
      Bali, Kalika",
    booktitle = "Proceedings of the 2023 Conference on Empirical Methods in Natural Language Processing",
    year = "2023"
}

@article{lin2022teaching,
  title={Teaching models to express their uncertainty in words},
  author={Lin, Stephanie and Hilton, Jacob and Evans, Owain},
  journal={arXiv preprint arXiv:2205.14334},
  year={2022}
}

@article{darrin2022rainproof,
  title={Rainproof: An umbrella to shield text generators from out-of-distribution data},
  author={Darrin, Maxime and Piantanida, Pablo and Colombo, Pierre},
  journal={arXiv preprint arXiv:2212.09171},
  year={2022}
}

@inproceedings{renout,
  title={Out-of-Distribution Detection and Selective Generation for Conditional Language Models},
  author={Ren, Jie and Luo, Jiaming and Zhao, Yao and Krishna, Kundan and Saleh, Mohammad and Lakshminarayanan, Balaji and Liu, Peter J},
  booktitle={The Eleventh International Conference on Learning Representations},
year={2023}
}

@inproceedings{takayama2019relevant,
  title={Relevant and informative response generation using pointwise mutual information},
  author={Takayama, Junya and Arase, Yuki},
  booktitle={Proceedings of the First Workshop on NLP for Conversational AI},
  pages={133--138},
  year={2019}
}

@article{fadeeva2024fact,
  title={Fact-checking the output of large language models via token-level uncertainty quantification},
  author={Fadeeva, Ekaterina and Rubashevskii, Aleksandr and Shelmanov, Artem and Petrakov, Sergey and Li, Haonan and Mubarak, Hamdy and Tsymbalov, Evgenii and Kuzmin, Gleb and Panchenko, Alexander and Baldwin, Timothy and others},
  journal={arXiv preprint arXiv:2403.04696},
  year={2024}
}

@article{lingenerating,
  title={Generating with Confidence: Uncertainty Quantification for Black-box Large Language Models},
  author={Lin, Zhen and Trivedi, Shubhendu and Sun, Jimeng},
  journal={Transactions on Machine Learning Research},
year={2023}
}

@article{savage2025large,
  title={Large language model uncertainty proxies: discrimination and calibration for medical diagnosis and treatment},
  author={Savage, Thomas and Wang, John and Gallo, Robert and Boukil, Abdessalem and Patel, Vishwesh and Safavi-Naini, Seyed Amir Ahmad and Soroush, Ali and Chen, Jonathan H},
  journal={Journal of the American Medical Informatics Association},
  volume={32},
  number={1},
  pages={139--149},
  year={2025},
  publisher={Oxford University Press}
}

@article{atf2025challenge,
  title={The challenge of uncertainty quantification of large language models in medicine},
  author={Atf, Zahra and Safavi-Naini, Seyed Amir Ahmad and Lewis, Peter R and Mahjoubfar, Aref and Naderi, Nariman and Savage, Thomas R and Soroush, Ali},
  journal={arXiv preprint arXiv:2504.05278},
  year={2025}
}

@article{wu2024uncertainty,
  title={Uncertainty estimation of large language models in medical question answering},
  author={Wu, Jiaxin and Yu, Yizhou and Zhou, Hong-Yu},
  journal={arXiv preprint arXiv:2407.08662},
  year={2024}
}

@article{qin2024enhancing,
  title={Enhancing healthcare llm trust with atypical presentations recalibration},
  author={Qin, Jeremy and Liu, Bang and Nguyen, Quoc Dinh},
  journal={arXiv preprint arXiv:2409.03225},
  year={2024}
}

@article{gao2025uncertainty,
  title={Uncertainty estimation in diagnosis generation from large language models: next-word probability is not pre-test probability},
  author={Gao, Yanjun and Myers, Skatje and Chen, Shan and Dligach, Dmitriy and Miller, Timothy and Bitterman, Danielle S and Chen, Guanhua and Mayampurath, Anoop and Churpek, Matthew M and Afshar, Majid},
  journal={JAMIA open},
  volume={8},
  number={1},
  pages={ooae154},
  year={2025},
  publisher={Oxford University Press}
}

@article{chen2024uncertainty,
  title={Uncertainty Quantification for Clinical Outcome Predictions with (Large) Language Models},
  author={Chen, Zizhang and Li, Peizhao and Dong, Xiaomeng and Hong, Pengyu},
  journal={arXiv preprint arXiv:2411.03497},
  year={2024}
}

@article{hu2024uncertainty,
  title={Uncertainty of thoughts: Uncertainty-aware planning enhances information seeking in llms},
  author={Hu, Zhiyuan and Liu, Chumin and Feng, Xidong and Zhao, Yilun and Ng, See-Kiong and Luu, Anh Tuan and He, Junxian and Koh, Pang Wei W and Hooi, Bryan},
  journal={Advances in Neural Information Processing Systems},
  volume={37},
  pages={24181--24215},
  year={2024}
}

@article{gu2024probabilistic,
  title={Probabilistic medical predictions of large language models},
  author={Gu, Bowen and Desai, Rishi J and Lin, Kueiyu Joshua and Yang, Jie},
  journal={npj Digital Medicine},
  volume={7},
  number={1},
  pages={367},
  year={2024},
  publisher={Nature Publishing Group UK London}
}

@article{omar2024benchmarking,
  title={Benchmarking the confidence of large language models in clinical questions},
  author={Omar, Mahmud and Agbareia, Reem and Glicksberg, Benjamin S and Nadkarni, Girish N and Klang, Eyal},
  journal={MedRxiv},
  pages={2024--08},
  year={2024},
  publisher={Cold Spring Harbor Laboratory Press}
}

@article{mehrtash2020confidence,
  title={Confidence calibration and predictive uncertainty estimation for deep medical image segmentation},
  author={Mehrtash, Alireza and Wells, William M and Tempany, Clare M and Abolmaesumi, Purang and Kapur, Tina},
  journal={IEEE transactions on medical imaging},
  volume={39},
  number={12},
  pages={3868--3878},
  year={2020},
  publisher={IEEE}
}

@article{zou2023review,
  title={A review of uncertainty estimation and its application in medical imaging},
  author={Zou, Ke and Chen, Zhihao and Yuan, Xuedong and Shen, Xiaojing and Wang, Meng and Fu, Huazhu},
  journal={Meta-Radiology},
  volume={1},
  number={1},
  pages={100003},
  year={2023},
  publisher={Elsevier}
}

@inproceedings{lohr2024towards,
  title={Towards aleatoric and epistemic uncertainty in medical image classification},
  author={L{\"o}hr, Timo and Ingrisch, Michael and H{\"u}llermeier, Eyke},
  booktitle={International Conference on Artificial Intelligence in Medicine},
  pages={145--155},
  year={2024},
  organization={Springer}
}

@inproceedings{abboud2024sparse,
  title={Sparse bayesian networks: Efficient uncertainty quantification in medical image analysis},
  author={Abboud, Zeinab and Lombaert, Herve and Kadoury, Samuel},
  booktitle={International Conference on Medical Image Computing and Computer-Assisted Intervention},
  pages={675--684},
  year={2024},
  organization={Springer}
}

@article{fansi2022ddxplus,
  title={Ddxplus: A new dataset for automatic medical diagnosis},
  author={Fansi Tchango, Arsene and Goel, Rishab and Wen, Zhi and Martel, Julien and Ghosn, Joumana},
  journal={Advances in neural information processing systems},
  volume={35},
  pages={31306--31318},
  year={2022}
}

@inproceedings{saley2024meditod,
  title={MediTOD: An English Dialogue Dataset for Medical History Taking with Comprehensive Annotations},
  author={Saley, Vishal and Saha, Goonjan and Das, Rocktim and Raghu, Dinesh and others},
  booktitle={Proceedings of the 2024 Conference on Empirical Methods in Natural Language Processing},
  pages={16843--16877},
  year={2024}
}

@article{jin2021medqa,
  title={What disease does this patient have? a large-scale open domain question answering dataset from medical exams},
  author={Jin, Di and Pan, Eileen and Oufattole, Nassim and Weng, Wei-Hung and Fang, Hanyi and Szolovits, Peter},
  journal={Applied Sciences},
  volume={11},
  number={14},
  pages={6421},
  year={2021},
  publisher={MDPI}
}

@article{lewis2020retrieval,
  title={Retrieval-augmented generation for knowledge-intensive nlp tasks},
  author={Lewis, Patrick and Perez, Ethan and Piktus, Aleksandra and Petroni, Fabio and Karpukhin, Vladimir and Goyal, Naman and K{\"u}ttler, Heinrich and Lewis, Mike and Yih, Wen-tau and Rockt{\"a}schel, Tim and others},
  journal={Advances in neural information processing systems},
  volume={33},
  pages={9459--9474},
  year={2020}
}

@inproceedings{chen-mueller-2024-quantifying,
    title = "Quantifying Uncertainty in Answers from any Language Model and Enhancing their Trustworthiness",
    author = "Chen, Jiuhai  and
      Mueller, Jonas",
    editor = "Ku, Lun-Wei  and
      Martins, Andre  and
      Srikumar, Vivek",
    booktitle = "Proceedings of the 62nd Annual Meeting of the Association for Computational Linguistics (Volume 1: Long Papers)",
    month = aug,
    year = "2024"
}

@inproceedings{van2022mutual,
  title={Mutual Information Alleviates Hallucinations in Abstractive Summarization},
  author={Van Der Poel, Liam and Cotterell, Ryan and Meister, Clara},
  booktitle={Proceedings of the 2022 Conference on Empirical Methods in Natural Language Processing},
  pages={5956--5965},
  year={2022}
}

@inproceedings{malininuncertainty,
  title={Uncertainty Estimation in Autoregressive Structured Prediction},
  author={Malinin, Andrey and Gales, Mark},
  booktitle={International Conference on Learning Representations},
  year={2021}
}

@misc{gpt-4.1,
    title = {Introducing GPT-4.1 in the API},
    author = {OpenAI},
    year = {2025},
    note = {\url{https://openai.com/index/gpt-4-1/}[accessed: 2025-08-25]}
}

@article{grattafiori2024llama,
  title={The llama 3 herd of models},
  author={Grattafiori, Aaron and Dubey, Abhimanyu and Jauhri, Abhinav and Pandey, Abhinav and Kadian, Abhishek and Al-Dahle, Ahmad and Letman, Aiesha and Mathur, Akhil and Schelten, Alan and Vaughan, Alex and others},
  journal={arXiv preprint arXiv:2407.21783},
  year={2024}
}

@article{robertson2009bm25,
  title={The probabilistic relevance framework: BM25 and beyond},
  author={Robertson, Stephen and Zaragoza, Hugo and others},
  journal={Foundations and Trends{\textregistered} in Information Retrieval},
  volume={3},
  number={4},
  pages={333--389},
  year={2009},
  publisher={Now Publishers, Inc.}
}

@article{ren2024healthcare,
  title={Healthcare agent: eliciting the power of large language models for medical consultation},
  author={Ren, Zhiyao and Zhan, Yibing and Yu, Baosheng and Ding, Liang and Xu, Pingbo and Tao, Dacheng},
  journal={npj Artificial Intelligence},
  volume={1},
  number={1},
  pages={24},
  year={2025},
  publisher={Nature Publishing Group UK London}
}

@inproceedings{xiongcan,
  title={Can LLMs Express Their Uncertainty? An Empirical Evaluation of Confidence Elicitation in LLMs},
  author={Xiong, Miao and Hu, Zhiyuan and Lu, Xinyang and LI, YIFEI and Fu, Jie and He, Junxian and Hooi, Bryan},
  booktitle={The Twelfth International Conference on Learning Representations},
  year = {2024}
}

@inproceedings{xiong2024benchmarking,
  title={Benchmarking retrieval-augmented generation for medicine},
  author={Xiong, Guangzhi and Jin, Qiao and Lu, Zhiyong and Zhang, Aidong},
  booktitle={Findings of the Association for Computational Linguistics ACL 2024},
  pages={6233--6251},
  year={2024}
}

@article{park2025deep,
  title={Deep Gaussian process with uncertainty estimation for microsatellite instability and immunotherapy response prediction from histology},
  author={Park, Sunho and Pettigrew, Morgan F and Cha, Yoon Jin and Kim, In-Ho and Kim, Minji and Banerjee, Imon and Barnfather, Isabel and Clemenceau, Jean R and Jang, Inyeop and Kim, Hyunki and others},
  journal={npj Digital Medicine},
  volume={8},
  number={1},
  pages={294},
  year={2025},
  publisher={Nature Publishing Group UK London}
}

@article{kang2021statistical,
  title={Statistical uncertainty quantification to augment clinical decision support: a first implementation in sleep medicine},
  author={Kang, Dae Y and DeYoung, Pamela N and Tantiongloc, Justin and Coleman, Todd P and Owens, Robert L},
  journal={NPJ digital medicine},
  volume={4},
  number={1},
  pages={142},
  year={2021},
  publisher={Nature Publishing Group UK London}
}

@article{kompa2021second,
  title={Second opinion needed: communicating uncertainty in medical machine learning},
  author={Kompa, Benjamin and Snoek, Jasper and Beam, Andrew L},
  journal={NPJ Digital Medicine},
  volume={4},
  number={1},
  pages={4},
  year={2021},
  publisher={Nature Publishing Group UK London}
}

@inproceedings{isenegger2019characterizing,
  title={Characterizing and quantifying diagnostic (un) certainty in medical reports through natural language processing},
  author={Isenegger, Kathleen and Dong, Yilan and Shang, Mengyuan and Furst, Jacob and Stan-Raicu, Daniela},
  booktitle={2019 International Conference on Computational Science and Computational Intelligence (CSCI)},
  pages={914--919},
  year={2019},
  organization={IEEE}
}

@article{peluso2024deep,
  title={Deep learning uncertainty quantification for clinical text classification},
  author={Peluso, Alina and Danciu, Ioana and Yoon, Hong-Jun and Yusof, Jamaludin Mohd and Bhattacharya, Tanmoy and Spannaus, Adam and Schaefferkoetter, Noah and Durbin, Eric B and Wu, Xiao-Cheng and Stroup, Antoinette and others},
  journal={Journal of biomedical informatics},
  volume={149},
  pages={104576},
  year={2024},
  publisher={Elsevier}
}

@inproceedings{khandokar2024towards,
  title={Towards precision diagnosis: Integrating lexical analysis and deep learning for uncertainty detection and quantification in clinical reports},
  author={Khandokar, Iftakhar and Farghaly, Omar and Kothari, Anai N and Deshpande, Priya},
  booktitle={2024 IEEE 37th International Symposium on Computer-Based Medical Systems (CBMS)},
  pages={267--272},
  year={2024},
  organization={IEEE}
}

@article{mukaka2012guide,
  title={A guide to appropriate use of correlation coefficient in medical research},
  author={Mukaka, Mavuto M},
  journal={Malawi medical journal},
  volume={24},
  number={3},
  pages={69--71},
  year={2012}
}

@inproceedings{davis2006relationship,
  title={The relationship between Precision-Recall and ROC curves},
  author={Davis, Jesse and Goadrich, Mark},
  booktitle={Proceedings of the 23rd international conference on Machine learning},
  pages={233--240},
  year={2006}
}

@article{hanley1982meaning,
  title={The meaning and use of the area under a receiver operating characteristic (ROC) curve.},
  author={Hanley, James A and McNeil, Barbara J},
  journal={Radiology},
  volume={143},
  number={1},
  pages={29--36},
  year={1982}
}

@article{ho2024novo,
  title={NoVo: Norm Voting off Hallucinations with Attention Heads in Large Language Models},
  author={Ho, Zheng Yi and Liang, Siyuan and Zhang, Sen and Zhan, Yibing and Tao, Dacheng},
  journal={arXiv preprint arXiv:2410.08970},
  year={2024}
}

@article{ying2026safebench,
  title={Safebench: A safety evaluation framework for multimodal large language models},
  author={Ying, Zonghao and Liu, Aishan and Liang, Siyuan and Huang, Lei and Guo, Jinyang and Zhou, Wenbo and Liu, Xianglong and Tao, Dacheng},
  journal={International Journal of Computer Vision},
  volume={134},
  number={1},
  pages={18},
  year={2026},
  publisher={Springer}
}

@article{liu2025agentsafe,
  title={AGENTSAFE: Benchmarking the Safety of Embodied Agents on Hazardous Instructions},
  author={Liu, Aishan and Ying, Zonghao and Wang, Le and Mu, Junjie and Guo, Jinyang and Wang, Jiakai and Ma, Yuqing and Liang, Siyuan and Zhang, Mingchuan and Liu, Xianglong and others},
  journal={arXiv preprint arXiv:2506.14697},
  year={2025}
}

@article{liang2025safemobile,
  title={SafeMobile: Chain-level Jailbreak Detection and Automated Evaluation for Multimodal Mobile Agents},
  author={Liang, Siyuan and Fang, Tianmeng and Liu, Zhe and Liu, Aishan and Xiao, Yan and He, Jinyuan and Chang, Ee-Chien and Cao, Xiaochun},
  journal={arXiv preprint arXiv:2507.00841},
  year={2025}
}

@article{liu2025roboview,
  title={RoboView-Bias: Benchmarking Visual Bias in Embodied Agents for Robotic Manipulation},
  author={Liu, Enguang and Liang, Siyuan and Lu, Liming and Zeng, Xiyu and Cao, Xiaochun and Liu, Aishan and Pang, Shuchao},
  journal={arXiv preprint arXiv:2509.22356},
  year={2025}
}

@article{chen2025mllms,
  title={Where MLLMs Attend and What They Rely On: Explaining Autoregressive Token Generation},
  author={Chen, Ruoyu and Guo, Xiaoqing and Liu, Kangwei and Liang, Si Yuan and Liu, Shiming and Zhang, Qunli and Zhang, Hua and Cao, Xiaochun},
  year={2025}
}

@article{liang2025explaining,
  title={Explaining multimodal LLMs via intra-modal token interactions},
  author={Liang, Jiawei and Chen, Ruoyu and Jiao, Xianghao and Liang, Siyuan and Liu, Shiming and Zhang, Qunli and Hu, Zheng and Cao, Xiaochun},
  journal={arXiv preprint arXiv:2509.22415},
  year={2025}
}

@article{ren2026dynamic,
  title={Dynamic Evaluation of Medical Large Language Models: A Survey Beyond Static Benchmarks},
  author={Ren, Zhiyao and Zhan, Yibing and Liang, Siyuan and Yu, Baosheng and Tao, Dacheng},
  journal={Authorea Preprints},
  year={2026},
  publisher={Authorea}
}

@article{ho2026understanding,
  title={Understanding Hallucinations in Large Visual and Language Models},
  author={Ho, Zheng Yi and Liang, Siyuan and Tao, Dacheng},
  journal={ACM Computing Surveys},
  volume={58},
  number={13},
  pages={1--36},
  year={2026},
  publisher={ACM New York, NY}
}

@inproceedings{zhang2026controllable,
  title={Controllable Contamination Detection for Reliable LLM Evaluation with Statistical Guarantees},
  author={Zhang, Zheng and Liu, Qi and Liang, Siyuan and Li, Ning and Hu, Zirui and Gao, Weibo and Li, Rui and Huang, Zhenya and Rutkowski, Leszek and Yu, Baosheng and others},
  booktitle={Proceedings of the 64th Annual Meeting of the Association for Computational Linguistics (Volume 1: Long Papers)},
  pages={30122--30143},
  year={2026}
}

\clearpage
\onecolumn
\begin{appendices}

\renewcommand{\thefigure}{A\arabic{figure}}
\renewcommand{\thetable}{A\arabic{table}}
\setcounter{figure}{0}
\setcounter{table}{0}
\renewcommand{\theequation}{A\arabic{equation}}
\setcounter{equation}{0}

\section{Common notations}
We list common notations in \Tref{tab:notion} for mathematical definitions. 

\begin{table*}[h]
\centering
\small
\caption{Common notations and descriptions.}
\label{tab:notion}
\begin{tabular}{l|l}
\toprule
\textbf{Notion} & \textbf{Description} \\
\midrule
$\mathbf{x}$ & Input sequence (patient information) \\
$\mathbf{y}$ & Output sequence (generated diagnosis/response) \\
$y_l$ & Token at position $l$ in the output sequence \\
$\mathbf{y}_{<l}$ & All tokens before position $l$ \\
$L$ & Length of generated sequence \\
$K$ & Number of sampled responses \\
$P(y_l | \mathbf{y}_{<l}, x)$ & Probability of token $y_l$ given previous tokens and input \\
$f(\cdot)$ & Large language model \\
$C$ & Confidence score \\
$\psi(\cdot)$ & Token-level transformation function \\
$\phi(\cdot)$ & Final transformation function \\
$w_l$ & Weight for token at position $l$ \\
$\bigoplus_{l=1}^L$ & Aggregation operation over tokens \\
$H(y_l | y_{<l}, x)$ & Entropy at token position $l$ \\
$N$ & Vocabulary size \\
$\mathbf{q}$ & Uniform distribution over vocabulary \\
$\alpha$ & Hyperparameter for Rényi divergence \\
$R_T(y_l, \mathbf{y}, \mathbf{x})$ & Token relevance function \\
$\tilde{R}_T(y_l, \mathbf{y}, \mathbf{x})$ & Normalized token relevance \\
$\mathcal{Y} = \{y_1, y_2, \ldots, y_K\}$ & Set of $K$ generated responses \\
$\mathbf{y}^{(k)}$ & The $k$-th sampled response \\
$\mathbf{y}_c$ & Most frequent/consistent response \\
$e(\cdot)$ & Embedding function \\
$\text{Cos}(\cdot, \cdot)$ & Cosine similarity function \\
$\mathcal{C}_n$ & Semantic cluster $n$ \\
$\tilde{P}_n(x)$ & Average probability of cluster $n$ \\
$\mathbf{S}$ & Similarity matrix \\
$\mathbf{L}$ & Graph Laplacian matrix \\
$\mathbf{D}$ & Diagonal degree matrix \\
$\lambda_k$ & $k$-th eigenvalue \\
$\mathbf{u}_k$ & $k$-th eigenvector \\
$\mathbf{v}_j$ & Embedding vector for response $j$ \\
$L_k$ & Length of sequence $y^{(k)}$ \\
$R_S(y^{(j)}, x)$ & Sentence relevance function \\
$I$ & Patient information \\
$d$ & Generated diagnosis \\
$\mathcal{D}$ & Medical corpus (PubMed, StatPearls, etc.) \\
$R$ & Retrieved medical passages from RAG \\
$S$ & Symptom profile (set of symptoms) \\
$\text{Prompt}$ & Prompt template \\
$\text{RAG}(d, \mathcal{D})$ & Retrieval-augmented generation function \\
\bottomrule
\end{tabular}
\end{table*}

\clearpage
\section{More Details of Benchmark}
\subsection{Details of Data Processing}
To better align with real-world medical consultation scenarios, we process the datasets by converting the model input format into doctor-patient dialogues or medical reports, and restructure the tasks as open-end decision making.

DDXPlus is a large-scale contains different diagnosis with ground truth pathology, symptoms and antecedents. It collect data in a structured format and provide categorical and multi-choice symptoms symptoms and antecedents for more detailed description. In our processing procedure, we convert structured format information into doctor-patient dialogue format. Using GPT-4.1, we convert symptom information in the dataset into doctor inquiries, while the symptoms are transformed into patient responses, converting binary and numerical information into patients' natural language expressions. The specific prompts are shown in \Fref{fig:ddxplus}. The model will generate diagnostic results based on the generated dialogue.

\begin{figure}[h]
\begin{tcolorbox}[title=Prompt for input conversion of DDXPlus]
\textbf{Prompt for doctor inquiry:}\\
You are playing the role of a doctor. Refer to the question \{question\}, ask the patient the question in a doctor's specialized form. Please do not change the original intent and do not include additional information and your response should contain only the question.\\
Question:\\
\hdashrule[0.5ex]{\linewidth}{1pt}{3mm 2pt}
\textbf{Prompt for patient answer:}\\
You are playing the role of a patient. Refer to the question \{question\} and the result {result}, answer the question in a patient's colloquial manner. Please do not include add any information that isn't mentioned and your response should contain only the answer.\\
Answer:
\end{tcolorbox}
\caption{Prompt for input conversion of DDXPlus.}
\label{fig:ddxplus}
\end{figure}

MediTOD comprises real-world doctor-patient dialogues with annotated complex relationships between dialogue content and corresponding clinical attributes. To select the effective dialogues, we retain the dialogue with intents of ``Inform", ``Inquire", and ``Diagnose", while ignore dialogue for ``Chit-chat" and ``salutations". MedQA contains profession multiple-choice question. The question background of MedQA is a comprehensive medical report and we remove the option in the question to convert the task from multiple choice question to open-end decision making task.

\subsection{Details of Evaluated Methods}
\subsubsection{Token-level Methods}
Average Sequence Probability (ASP)~\cite{huang2025look} computes the arithmetic mean of token probabilities across the generated sequence:
\begin{equation}
C_{\text{ASP}} = \frac{1}{L} \sum_{l=1}^{L} P(y_l \mid \mathbf{y}_{<l}, \mathbf{x})
\end{equation}
Maximum Sequence Probability (MSP)~\cite{huang2025look} estimates the sequence-level confidence by selecting the highest token probability within the generated sequence:
\begin{equation}
C_{\text{MSP}} = \max_{l=1,...,L} P(y_l \mid \mathbf{y}_{<l}, \mathbf{x})
\end{equation}
Additional metrics analyze the probability distribution characteristics. Perplexity~\cite{renout} measures the exponential of the average negative log-probability: 
\begin{equation}
   C_{\text{Perp}} = \exp\left\{-\frac{1}{L} \sum_{l=1}^{L} \log P(y_l | \mathbf{y}_{<l}, \mathbf{x})\right\}
\end{equation}
The Entropy~\cite{manakul2023selfcheckgpt} computes the average entropy across all token positions in the generated sequence:
\begin{equation}
    C_{\mathcal{H}} = \frac{1}{L} \sum_{l=1}^{L} \mathcal{H}(y_l | y_{<l}, \mathbf{x})
\end{equation}
where $\mathcal{H}(y_l | y_{<l}, \mathbf{x})$ represents the entropy of the token distribution at position $l$.

\citet{takayama2019relevant} propose utilize Pointwise Mutual Information (PMI) between conditioned and unconditioned input to measure confidence:
\begin{equation}
    C_{\text{PMI}} = \frac{1}{L} \sum_{l=1}^{L} \log \frac{P(y_l | y_{<l})}{P(y_l | y_{<l}, \mathbf{x})}
\end{equation}

\citet{van2022mutual} propose a modification called Conditional Pointwise Mutual Information (CPMI) to only evaluate those token have entropy above a threshold:
\begin{align}
C_{\text{CPMI}} &= \frac{1}{L} \sum_{l=1}^{L} \log P(y_l | y_{<l}, \mathbf{x}) \\
&\quad + \frac{\lambda}{L} \sum_{l: \mathcal{H}(y_l | y_{<l}, \mathbf{x}) \geq \tau} \log P(y_l | y_{<l})
\end{align}
where $\lambda$ is a changeable parameter. 

\citet{darrin2022rainproof} propose to use Rényi Divergence and Fisher-Rao distance to calculate the divergence between the distribution of each token and the uniform distribution. 
For $N$ is the number of tokens in vocabulary and $\mathbf{q}$ is a uniform distribution over the vocabulary, the Renyi divergences is:
\begin{equation}
C_{\text{RD}} = \frac{1}{L} \sum_{l=1}^{L} \frac{1}{\alpha - 1} \log \sum_{i=1}^{N} \frac{P(y_l | y_{<l}, \mathbf{x})^{\alpha}}{\mathbf{q}_i^{\alpha-1}} 
\end{equation}
while the Fisher-Rao is:
\begin{equation}
    C_{\text{FRD}} = \frac{1}{L} \sum_{l=1}^{L} \frac{2}{\pi} \arccos \sum_{i=1}^{N} \sqrt{P(y_l | y_{<l}, \mathbf{x}) \cdot \mathbf{q}_i}
\end{equation}

TokenSAR~\cite{duan2024shifting} first proposes that the contribution of different tokens are unequal. It utilize a token relevance function to calculate the importance of each token:
\begin{equation}
    R_T(y_l, \mathbf{y}, \mathbf{x}) = 1 - g(\mathbf{x} \cup \mathbf{y}, \mathbf{x} \cup \mathbf{y} \setminus \{y_l\})
\end{equation}
where $g(\cdot,\cdot)$ measures the semantic similarity between two sequences. The reweighted entropy is then computed as:
\begin{equation}
\label{eq:tokensar}
    C_{\text{TokenSAR}} = -\sum_{l=1}^{L} \tilde{R}_T(y_l, \mathbf{y}, \mathbf{x}) \log P(y_l | y_{<l}, \mathbf{x}) 
\end{equation}
where $\tilde{R}_T(y_l, \mathbf{y}, \mathbf{x}) = \frac{R_T(y_l, \mathbf{y}, \mathbf{x})}{\sum_{i=1}^{L} R_T(y_i, \mathbf{y}, \mathbf{x})}$ represents the normalized token relevance weights.

Claim Conditioned Probability (CCP)~\cite{fadeeva2024fact} quantifies uncertainty by assessing how token substitutions affect semantic consistency. For each token position $j$, the method replaces token $y_j$ with alternative candidates $y_j^k$ sampled from the top-$k$ predictions of the model's output distribution. A Natural Language Inference (NLI) model then evaluates the semantic relationship between the original sequence and each perturbed variant. The CCP score is computed as:
\begin{equation}
    C_{\text{CCP}} = \frac{\sum_{k: \text{NLI}(y_j^k, y_j) = \text{`entail'}} P(y_j^k | y_{<j}, \mathbf{x})}{\sum_{k: \text{NLI}(y_j^k, y_j) \in \{\text{`entail'}, \text{`contra'}\}} P(y_j^k | y_{<j}, \mathbf{x})} 
\end{equation}
where $\text{NLI}(y_j^k, y_j) = \text{`entail'}$ denotes the NLI model predicts an entailment relation and $\text{NLI}(y_j^k, y_j) = \text{`contra'}$ denotes that the NLI model classifies relationships as contradiction.

\subsubsection{Consistency-level Methods}
Percentage of Consistency (PoC)~\cite{manakul2023selfcheckgpt} evaluates confidence by calculating the proportion of responses that match the most frequent response:
\begin{equation}
    C_{\text{PoC}} = \frac{\sum_{i=1}^K \mathbb{I}(\mathbf{y}_i = \mathbf{y}_c)}{K}
\end{equation}
where $\mathbf{y}_c = \arg\max_{\mathbf{y} \in \mathcal{Y}} \sum_{j=1}^K \mathbb{I}(\mathbf{y}_j = \mathbf{y})$.

Semantic Similarity (SemSim)~\cite{chen-mueller-2024-quantifying} measures confidence by computing the average cosine similarity between embeddings of all response pairs. It can be represented as:
\begin{equation}
    C_{\text{SemSim}} = \frac{1}{K(K-1)} \sum_{i=1}^K \sum_{j \neq i} \text{Cos}(e(\mathbf{y}_i), e(\mathbf{y}_j))
\end{equation}
where $\text{Cos}(\cdot,\cdot)$ denotes the cosine similarity function and $e(\cdot)$ represents the embedding process that maps sequences to vector representations.  

Lexical Similarity~\cite{manakul2023selfcheckgpt} measures confidence through lexical overlap between K sampled responses using n-gram metric (\eg, ROUGE and BLEU):
\begin{equation}
    C_{\text{LexSim}} = \frac{1}{K(K-1)} \sum_{i=1}^{K} \sum_{j \neq i} \text{LexicalSim}(y^{(i)}, y^{(j)})
\end{equation}
where LexicalSim computes n-gram overlap between responses $y^{(i)}$ and $y^{(j)}$. In our setting, we utilize ROUGE-L as the lexical similarity metric.

Number of Semantic Sets (NumSet)~\cite{lingenerating} clusters semantically similar responses into non-overlapping groups using NLI-based entailment relations. The confidence calculate the number of cluster in the response:
\begin{equation}
    C_{\text{NumSet}} = 1 - \frac{\text{NumClusters}}{K}
\end{equation}
where responses $y^{(i)}$ and $y^{(i)}$ are clustered together if $\hat{p}_{\text{entail}}(y^{(i)}, y^{(j)}) > \hat{p}_{\text{contra}}(y^{(i)}, y^{(j)})$ and $\hat{p}_{\text{entail}}(y^{(j)}, y^{(i)}) > \hat{p}_{\text{contra}}(y^{(j)}, y^{(i)})$.

Sum of Eigenvalues of the Graph Laplacian (EigV)~\cite{lingenerating} constructs a similarity graph from K responses and analyzes eigenvalues of the normalized Laplacian matrix to quantify semantic diversity. For a similarity matrix $S$, the Laplacian for $S$ is formula as:
\begin{equation}
    L = I - D^{-\frac{1}{2}} S D^{-\frac{1}{2}}
\end{equation}
where $D$ is a diagonal matrix and $\lambda_k$ are the eigenvalues of matrix $L$. The confidence of EigV can be represented as:
\begin{equation}
    C_{\text{EigV}} = 1 - \sum_{k=1}^{K} \max(0, 1 - \lambda_k)
\end{equation}

Furthermore, Degree Matrix (Deg)~\cite{lingenerating} extract confidence from the previous metioned diagonal matrix $D$ and the confidence score is estimated by:
\begin{equation}
    C_{\text{Deg}} = \frac{\text{trace}(D)}{K^2}
\end{equation}

Eccentricity (Ecc)~\cite{lingenerating} quantifies confidence by measuring distance from the centroid in embedding space derived from the similarity graph. This method leverages the eigenvectors $\mathbf{u}_1, \cdots, \mathbf{u}_k$ corresponding to the k smallest eigenvalues of the graph Laplacian to construct informative embeddings $v_j = [\mathbf{u}_{1,j}, \ldots, \mathbf{u}_{k,j}]$ for each response $\mathbf{y}_j$. The uncertainty score is calculated as the average distance from the centroid in this embedding space: 
\begin{equation}
    C_{\text{Ecc}} = 1 - \|[\tilde{v}_1^T, \ldots, \tilde{v}_K^T]\|_2
\end{equation}
where $\tilde{\mathbf{v}}_j = \mathbf{v}_j - \frac{1}{K} \sum_{\ell=1}^{K} \mathbf{v}_\ell$.

On the other method, several approaches integrate token probability data into consistency measures. Monte Carlo Sequence Entropy (MC-SE)~\cite{kuhn2023semantic} calculates entropy at the sequence level by averaging the negative log-probabilities across multiple generated sequences: 
\begin{equation}
    C_{\text{MC-SE}} = -\frac{1}{K} \sum_{k=1}^{K} \log P(y^{(k)} | \mathbf{x})
\end{equation}
where $P(\mathbf{y}^{(k)} | \mathbf{x}) = \prod_{l=1}^{L_k} P(y_l^{(k)} | \mathbf{y}_{<l}^{(k)}, \mathbf{x})$ represents the probability of the $k$-th generated sequence. To obtain a more reliable uncertainty measure, the sequence probabilities can be length-normalized~\cite{malininuncertainty}:
\begin{equation}
    \hat{C}_{\text{MC-SE}} = -\frac{1}{K} \sum_{k=1}^{K} \frac{1}{L_k} \log P(\mathbf{y}^{(k)} | \mathbf{x})
\end{equation}
where $L_k$ is the length of sequence $\mathbf{y}^{(k)}$.

Semantic Entropy (SE)~\cite{kuhn2023semantic} specifically targets the influence of semantically equivalent expressions on entropy calculations. It clusters generated responses into semantically similar groups $\mathcal{C}_n, n = 1, 2, \ldots, N$ and averages the sequence probabilities within each cluster. The entropy calculated over these semantic clusters is formulated as:
\begin{equation}
    C_{\text{SE}} = -\sum_{n=1}^{N} \frac{|C_n|}{K} \log \tilde{P}_n(\mathbf{x})
\end{equation}
where $\tilde{P}_n(\mathbf{x}) = \sum_{\mathbf{y} \in \mathcal{C}_n} P(\mathbf{y} | \mathbf{x})$ represents the average probability of cluster $n$.

For SentenceSAR~\cite{duan2024shifting}, it reweigh the importance of each sentence. Through the sentence relevance function:
\begin{equation}
    R_S(y^{(j)}, \mathbf{x}) = \sum_{k \neq j} g(y^{(j)}, y^{(k)}) P(y^{(k)} | \mathbf{x})
\end{equation}
it increase the probability of sentences that are more relevant than others:
\begin{equation}
    U_{\text{SentSAR}}(\mathbf{x}) = \frac{1}{K} \sum_{k=1}^{K} \left( \log P(y^{(k)} | \mathbf{x}) + \frac{1}{t} R_S(y^{(k)}, \mathbf{x}) \right)
\end{equation}
where $t$ is the hyperparameter.

By changing the probability $P(\mathbf{y} \mid \mathbf{x})$ with TokenSAR in Eq.~\ref{eq:tokensar}, we can obtain SAR method that combine both TokenSAR and SentenceSAR.

\subsubsection{Self-verbalized Methods}
We utilize three prompt strategies for Confidence Elicitation method, including vanilla prompt, chain-of-through prompt, and Tok-k prompt. The prompt details are shown in \Fref{fig:vanilla_prompt}, \Fref{fig:cot_prompt}, and \Fref{fig:topk_prompt}, respectively.

\begin{figure}[h]
    \begin{tcolorbox}[title=Prompt for vanilla CE]
Your task is to rate the confidence of the proposed answer on a score from 0 to 100, where 0 represents definitely not confidence and 100 represents definitely confidence. Note: The confidence indicates how likely you think your answer is true. \\
Please, only answer with your score in between square brackets (ex. [50]).\\
===========\\
Scenario: {dialogue}\\
Answer: {answer}
    \end{tcolorbox}
    \caption{Prompt for vanilla CE.}
    \label{fig:vanilla_prompt}
\end{figure}

\begin{figure}[h]
    \begin{tcolorbox}[title=Prompt for CoT CE]
Your task is to rate the confidence of the proposed answer on a score from 0 to 100, where 0 represents definitely not confidence and 100 represents definitely confidence. Note: The confidence indicates how likely you think your answer is true. \\
Provide your explanation first and show your confidence in between square brackets (ex. [50]). The answer format is:\\
Explanation:\\
Confidence:\\
===========\\
Scenario: {dialogue}\\
Answer: {answer}
    \end{tcolorbox}
    \caption{Prompt for CoT CE.}
    \label{fig:cot_prompt}
\end{figure}

\begin{figure}[h]
    \begin{tcolorbox}[title=Prompt for Top-k CE]
Your task is to rate the uncertainty of the proposed answer on a score from 0 to 100, where 0 represents definitely uncertain and 100 represents definitely certain. Note: The confidence indicates how likely you think your answer is true.\\
Provide your 5 best guess of the confidence and answer with your score in between square brackets (ex. [50]). For example:\\
G1: \\
...\\
G5: \\
===========\\
Scenario: {dialogue}\\
Answer: {answer}\\
Now, please provide your confidence.
    \end{tcolorbox}
    \caption{Prompt for Top-k CE.}
    \label{fig:topk_prompt}
\end{figure}

\section{Details of MedConf}
In the diagnostic generation step, given patient information, the model generate diagnosis with diagnosis prompt $\text{Prompt}_d$. The details of $\text{Prompt}_d$ are shown in \Fref{fig:diagnosis_prompt}.

\begin{figure}[h]
\begin{tcolorbox}[title=Details of diagnosis prompt]
Read the medical information below and determine the final diagnosis. Provide a  specific diagnosis for the case and label your Answer in square brackets. \\
===========\\
Medical information: \{inquiry\}
\end{tcolorbox}
\caption{Details of diagnosis prompt.}
\label{fig:diagnosis_prompt}
\end{figure}

Following this, RAG system identify and retrieve the relevant medical passages in database associated with the generated diagnosis. The retrieve results are summarize to a JSON structured symptom profile with symptom profile generation prompt $\text{Prompt}_s$. The details of $\text{Prompt}_s$ are shown in \Fref{fig:profile_prompt}.

\begin{figure}[h]
\tiny
\begin{tcolorbox}[title=Details of symptom profile generation prompt]
You are a helpful medical expert, and your task is to generate structured diagnostic criteria for the following condition using the relevant documents. Each criterion should include a description and importance level and output in JSON format.\\
Condition: \{diagnosis\}\\
Relevant documents: \{content\}\\

Now, Please output the diagnostic criteria in JSON format as follow:\\
\{\\
    ``id": Criterion ID, e.g., 1,\\
    ``description": "Specific diagnostic criterion description",\\
    ``importance": "strong/moderate/weak"\\
\}\\

Importance definitions:\\
- strong: Core diagnostic criteria, absence severely impacts diagnosis\\
- moderate: Important supportive criteria, helps confirm diagnosis\\
- weak: Auxiliary criteria, increases diagnostic confidence but not essential
\end{tcolorbox}
\caption{Details of symptom profile generation prompt.}
\label{fig:profile_prompt}
\end{figure}

Finally, the confidence generation step utilize the confidence generation prompt $\text{Prompt}_c$ to first analyze the supportive, missing, and contradictory relationship between patient information and symptom profile and then estimate confidence score. The details of $\text{Prompt}_c$ are shown in \Fref{fig:confidence_prompt}.

\begin{figure}[h]
\tiny
\begin{tcolorbox}[title=Details of confidence generation prompt]
You are a specialist with extensive medical knowledge. Your task is evaluate how well each diagnostic criterion is met based on the following patient-doctor conversation and then provide the confidence score. The confidence score is from 0 to 100, where 0 represents definitely not confidence and 100 represents definitely confidence. Note: The confidence indicates how likely you think your answer is true. \\
Diagnostic Criteria: \{criteria\}\\

Patient Information: \{inform\}\\

First, you need to analyze each criterion and generate evaluation results in the following JSON format:\\
\{\\
    ``criteria evaluation": [\\
        \{\\
            ``id": Criterion ID,\\
            ``description": Criterion description,\\
            ``importance": strong/moderate/weak,\\
            ``support level": supported/contradicted/missing,\\
            ``evidence": Quoted supporting/contradicting evidence from conversation, or null if none\\
        \}\\
    ],\\
    ``summary": \{\\
        ``supported criteria": \{\\
            ``supported strong": [List all supported strong-level criteria],\\
            ``supported moderate": [List all supported moderate-level criteria],\\
            ``supported weak": [List all supported weak-level criteria]\\
        \},\\
        ``missing criteria": \{\
            ``missing strong": [List all missing strong-level criteria],\\
            ``missing moderate": [List all missing moderate-level criteria],\\
            ``missing weak": [List all missing weak-level criteria]\\
        \}\\
        ``contradicted criteria": \{\\
            ``contradicted strong": [List all contradicted strong-level criteria],\\
            ``contradicted moderate": [List all contradicted moderate-level criteria],\\
            ``contradicted weak": [List all contradicted weak-level criteria]\\
        \},\\
        ``confidence": \{\\
            reasoning: Provide your reasons for the confidence score.\\
            confidence score: Please provide the confidence score between double angle brackets\\
        \}
    \}
\}
\end{tcolorbox}
\caption{Details of confidence generation prompt.}
\label{fig:confidence_prompt}
\end{figure}

\clearpage
\section{Case Study}
\subsection{Case Study of Existing Methods}
In this section, we provide case study for token-level methods, consistency-level methods, and self-verbalized methods. In our example, the model input is ``\textit{I have been having stomach pain, feeling nauseous, and I have a slight fever."}.

In this example, the model prediction is ``\textit{appendicitis"} and the probability for each token are $\textit{``append} = 0.3204, \textit{icit} = 0.9722, \textit{is}= 0.9999$". So, for token-level method ASP that calculate the mean of the generated token, the confidence result is 76.42. Furthermore, when the generate temperature is 0.5 and we generate 15 answers, there are 10 answers are appendicitis, 3 answers are acute gastroenteritis, and 2 answers are viral gastroenteritis. When we apply consistency-level methods PoC that calculate the proportion of the most frequent response, the confidence is 66.67. Finally, when we utilize self-verbalized method CoT CE to obtain the result, the model output are shown in \Fref{fig:case_COTCE}. The model provide an anlysis first and then evaluate the confidence score as 60.

\begin{figure}[h]
\begin{tcolorbox}[title=Details of CoT CE in case study]
Appendicitis is a condition where the appendix, a small organ attached to the large intestine, becomes inflamed.  The symptoms of appendicitis can vary, but common ones include abdominal pain, nausea, vomiting, and fever.  Given the symptoms you described, such as stomach pain, feeling nauseous, and a slight fever, it is possible that you may have appendicitis. Given the information provided, it is difficult to make a definitive diagnosis without further evaluation.  However, based on the symptoms described, I would rate the uncertainty of the proposed answer as 60.
\end{tcolorbox}
\caption{Details of CoT CE in case study.}
\label{fig:case_COTCE}
\end{figure}

\subsection{Case Study of MedConf}
\subsubsection{Diagnostic Generation}
In this example, the patient describes their symptoms to the doctor to seek a diagnosis. Their patient information $I$ is shown in \Fref{fig:case_patient_information}. Due to the patient's limited capabilities, they often can only provide incomplete and brief descriptions.

\begin{figure}[h]
\begin{tcolorbox}[title=Details of patient information in case study]
I have been having stomach pain, feeling nauseous, and I have a slight fever.
\end{tcolorbox}
\caption{Details of patient information in case study.}
\label{fig:case_patient_information}
\end{figure}

Subsequently, patient information $I$ is used to generate diagnostic assessment $d$ through the guidance of the diagnosis prompt $\text{Prompt}_d$. The generated result $d$ is shown in \Fref{fig:case_diagnostic_result}.

\begin{figure}[h]
\begin{tcolorbox}[title=Details of diagnostic result in case study]
I have been having stomach pain, feeling nauseous, and I have a slight fever.
\end{tcolorbox}
\caption{Details of diagnostic result in case study.}
\label{fig:case_diagnostic_result}
\end{figure}

\subsubsection{Symptom Generation}
In this process, LLMs generate a symptom profile for the diagnostic results, which serves as the support knowledge for subsequent confidence evaluation. In this example, we utilize RAG to obtain the most relevant passages in the medical corpus $\mathcal{D}$ of the keyword ``\textit{appendicitis"}. An example of RAG results $R$ is shown in \Fref{fig:case_RAG_result}.

\begin{figure}[h]
\tiny
\begin{tcolorbox}[title=Examples of RAG result in case study]
Document *0* (Title: Appendicitis) Epidemiology Appendicitis is most common between the ages of 5 and 40. In 2013, it resulted in 72,000 deaths globally, down from 88,000 in 1990. In the United States, there were nearly 293,000 hospitalizations involving appendicitis in 2010. Appendicitis is one of the most frequent diagnoses for emergency department visits resulting in hospitalization among children ages 5–17 years in the United States. See also Deaths from appendicitis Evan O'Neill Kane Leonid Rogozov References External links CT of the abdomen showing acute appendicitis Appendicitis, history, diagnosis and treatment by Surgeons Net Education Appendicitis: Acute Abdomen and Surgical Gastroenterology from the Merck Manual Professional (content last modified September 2007) Appendicitis – Symptoms Causes and Treatment at Health N Surgery
\end{tcolorbox}
\caption{Examples of RAG result in case study.}
\label{fig:case_RAG_result}
\end{figure}

Following the RAG step, the LLM generates symptom profile $S$ based on the RAG result $R$, the diagnostic result $d$, and the symptom profile generation prompt $\text{Prompt}_s$. The symptom profile for this example is shown in \Fref{fig:case_symptom_profile}.

\begin{figure}[h]
\tiny
\begin{tcolorbox}[title=Details of symptom profile in case study]
[\\    \{\\        ``id'': 1,\\        ``description'': ``Presence of right lower quadrant abdominal pain, which is often the first symptom of appendicitis.'',\\        ``importance'': ``strong''\\    \},\\    \{\\        ``id'': 2,\\        ``description'': ``Nausea and vomiting following the onset of abdominal pain, which is commonly reported by patients with appendicitis.'',\\        ``importance'': ``strong''\\    \},\\    \{\\        ``id'': 3,\\        ``description'': ``Fever, typically low-grade, which may accompany the other symptoms of appendicitis.'',\\        ``importance'': ``moderate''\\    \},\\    \{\\        ``id'': 4,\\        ``description'': ``Rebound tenderness or guarding in the right lower quadrant during physical examination, indicating irritation of the peritoneum.'',\\        ``importance'': ``strong''\\    \},\\    \{\\        ``id'': 5,\\        ``description'': ``Elevated white blood cell count (leukocytosis), which is a common laboratory finding in patients with appendicitis.'',\\        ``importance'': ``moderate''\\    \},\\    \{\\        ``id'': 6,\\        ``description'': ``Use of imaging studies such as ultrasound or CT scan to visualize the appendix and confirm the diagnosis of appendicitis.'',\\        ``importance'': ``moderate''\\    \},\\    \{\\        ``id'': 7,\\        ``description'': ``History of previous episodes of abdominal pain or gastrointestinal issues, which may suggest a recurrent appendicitis.'',\\        ``importance'': ``weak''\\    \},\\    \{\\        ``id'': 8,\\        ``description'': ``Application of scoring systems like the Alvarado score or Appendicitis Inflammatory Response (AIR) score to assist in diagnosis.'',\\        ``importance'': ``weak''\\    \}\\]
\end{tcolorbox}
\caption{Details of symptom profile in case study.}
\label{fig:case_symptom_profile}
\end{figure}

\subsubsection{Confidence Generation}
In this process, the LLMs evaluate the relationship between patient information $I$ and symptom profile $S$ and then utilize it as evidence to inference the confidence score of current generated diagnosis. The response of the final confidence is shown in \Fref{fig:case_confidence_generation}.

\begin{figure}[h]
\tiny
\begin{tcolorbox}[title=Details of confidence generation in case study]
\{\\    ``criteria\_evaluation": [\\        \{\\           ``id": 1,\\            ``description": ``Presence of right lower quadrant abdominal pain, which is often the first symptom of appendicitis.",\\            ``importance": ``strong",\\            ``support\_level":``"missing",\\            ``evidence": null\\        \},\\        \{\\            ``id": 2,\\            ``description": ``Nausea and vomiting following the onset of abdominal pain, which is commonly reported by patients with appendicitis.",\\            ``importance": ``strong",\\            ``support\_level": ``supported",\\            ``evidence": ``I have been feeling nauseous"\\        \},\\        \{\\            ``id": 3,\\            ``description": ``Fever, typically low-grade, which may accompany the other symptoms of appendicitis.",\\            ``importance": ``moderate",\\            ``support\_level": ``supported",\\            ``evidence": ``I have a slight fever"\\        \},\\        \{\\            ``id": 4,\\            ``description": ``Rebound tenderness or guarding in the right lower quadrant during physical examination, indicating irritation of the peritoneum.",\\            ``importance": ``strong",\\            ``support\_level": ``missing",\\            ``evidence": null\\        \},\\        \{\\            ``id": 5,\\            ``description": ``Elevated white blood cell count (leukocytosis), which is a common laboratory finding in patients with appendicitis.",\\            ``importance": ``moderate",\\            ``support\_level": ``missing",\\            ``evidence": null\\        \},\\        \{\\            ``id": 6,\\            ``description": ``Use of imaging studies such as ultrasound or CT scan to visualize the appendix and confirm the diagnosis of appendicitis.",\\            ``importance": ``moderate",\\            ``support\_level": ``missing",\\            ``evidence": null\\        \},\\        \{\\            ``id": 7,\\            ``description": ``History of previous episodes of abdominal pain or gastrointestinal issues, which may suggest a recurrent appendicitis.",\\            ``importance": ``weak",\\            ``support\_level": ``missing",\\            ``evidence": null\\        \},\\        \{\\            ``id": 8,\\            ``description": ``Application of scoring systems like the Alvarado score or Appendicitis Inflammatory Response (AIR) score to assist in diagnosis.",\\            ``importance": ``weak",\\            ``support\_level": ``missing",\\            ``evidence": null\\        \}\\    ],\\    ``summary": \{\\        ``supported\_criteria": \{\\            ``supported\_strong": [2],\\            ``supported\_moderate": [3],\\            ``supported\_weak": []\\        \},\\        ``missing\_criteria": \{\\            ``missing\_strong": [1, 4],\\            ``missing\_moderate": [5, 6],\\            ``missing\_weak": [7, 8]\\        \},\\        ``contradicted\_criteria": \{\\            ``contradicted\_strong": [],\\            ``contradicted\_moderate": [],\\            ``contradicted\_weak": []\\        \},\\        ``confidence": \{\\            ``reasoning": ``The patient has reported nausea and a slight fever, which supports two of the moderate and strong criteria. However, the absence of specific evidence for right lower quadrant pain and other criteria leads to a lower confidence score.",\\            ``confidence\_score": ``$<$<$30$>$>$"\\        \}\\    \}\}
\end{tcolorbox}
\caption{Details of confidence generation in case study.}
\label{fig:case_confidence_generation}
\end{figure}


\end{appendices}

\end{document}